\documentclass[lettersize, journal]{IEEEtran}

\ifCLASSINFOpdf
\else
\fi

\usepackage{url}
\usepackage{caption}
\usepackage{subcaption}
\usepackage{graphicx}
\usepackage{multicol}
\usepackage{multirow}
\usepackage{algorithm,algpseudocode}
\usepackage{algpseudocode}
\usepackage{zebra-goodies}
\usepackage{placeins} 
\usepackage{hyperref}

\usepackage{amsmath}
\usepackage{amssymb}
\usepackage{gensymb}
\usepackage{xltabular}
\usepackage{booktabs}
\usepackage{float}
\RequirePackage{doi}
\usepackage[numbers,sort&compress,square]{natbib}
\usepackage{tabularx}
\usepackage{array}
\usepackage[export]{adjustbox}
\usepackage{flushend}

\begin{document}
%
\title{On the Assessment of Sensitivity of Autonomous Vehicle Perception}
%
%
%

\author{Apostol Vassilev, Munawar Hasan, Edward Griffor, Honglan Jin, Pavel Piliptchak, Mahima Arora, Thoshitha Gamage
        
\thanks{All authors are with NIST or performed the work while at NIST}
\thanks{Manuscript received TBD; revised TBD.}}

%
%

\markboth{Journal of \LaTeX\ Class Files,~Vol.~14, No.~8, June~2025}%
{Shell \MakeLowercase{\textit{et al.}}: Bare Demo of IEEEtran.cls for IEEE Journals}
%



\maketitle
\begin{abstract}
The viability of automated driving is heavily dependent on the performance of perception systems to provide real-time accurate and reliable information for robust decision-making and maneuvers. These systems must perform reliably not only under ideal conditions, but also when challenged by natural and adversarial driving factors. These include inclement weather conditions and the presence of occluded objects. Both of these types of interference can lead to perception errors and delays in detection and classification. Hence, it is essential to assess the robustness of the perception systems of automated vehicles (AVs) and explore strategies for making perception more reliable.  The paper approaches this problem by evaluating perception performance using predictive sensitivity quantification based on an ensemble of models, capturing model disagreement and inference variability across multiple models, under adverse driving scenarios in both simulated environments and real-world conditions. A notional architecture for assessing perception performance is proposed that comprehends multiple input sources, including ROS-based interface to AI models, and extensible AI architecture that provides detection and classification as outputs along with  predictive sensitivity and post processing. Inputs can include videos of adversarial examples from real vehicle data and the CARLA simulation platform. 
A perception assessment criterion is developed based on an AV's stopping distance at a stop sign on varying road surfaces, such as dry and wet asphalt, and vehicle speed. Five state-of-the-art computer vision models are used, including YOLO (v8-v9), DEtection TRansformer (DETR50, DETR101), Real-Time DEtection TRansformer (RT-DETR)\footnote{Certain equipment, instruments, software, or materials, commercial or non-commercial, are identified in this paper in order to specify the experimental procedure adequately. Such identification does not imply recommendation or endorsement of any product or service by NIST, nor does it imply that the materials or equipment identified are necessarily the best available for the purpose.} in our experiments.
Diminished lighting conditions, e.g., resulting from the presence of fog and low sun altitude, are seen to have the greatest impact on the performance of the perception models. Additionally, adversarial road conditions such as occlusions of roadway objects increase perception sensitivity and model performance drops when faced with a combination of adversarial road conditions and inclement weather conditions. Also, it is demonstrated that the greater the distance to a roadway object, the greater the impact on perception performance, hence diminished perception robustness.   \end{abstract}

\begin{IEEEkeywords}
autonomous vehicle, deep ensemble, perception robustness, perception sensitivity,  adversarial conditions, safety quadrant
\end{IEEEkeywords}

\IEEEpeerreviewmaketitle

\section{Introduction}

As autonomous vehicles (AVs) approach real-world adoption, their ability to safely navigate complex environments hinges on the robustness of their perception systems, and remains a significant challenge \cite{Rasmus_Safety, Cummins2024}.
 Leading AV developers agree that one of the main challenges in the self-driving domain is the AV handling complex real-world scenarios along with complex edge cases \cite{cruise_challenges_AVs, waymo_challenges_AVs}. Another major challenge is to accurately evaluate and validate AV system's performance \cite{waymo_challenges_AVs}. Performing overall system validation begins with assuring individual components, such as AI perception, performance, followed by full system validation. These perception systems, powered by AI, must accurately detect and classify objects in the surrounding environment to make safe and informed driving decisions \cite{khan_challenges, Moloy2024}. However, AI models demonstrate inconsistent prediction behavior even with small variations in input. Such predictions can lead to critical errors, compromising the very foundation upon which the AV revolution is built \cite{Cummins2024-ROOT}. For example, false positives and false negatives can lead to unnecessary or phantom braking \cite{Nhtsa_investigation_url} or, more critically, a failure to react to actual hazards. Localization errors can result in collisions due to misjudged distances or unsafe maneuvers based on automated driving systems (ADS) localization estimates \cite{patel_localizationErrs}. Tracking errors, particularly under more complex driving conditions like intersections, can cause the vehicle to misinterpret the movements of other road users, leading to potentially devastating collisions. In addition to the challenges of object detection, misclassification of objects, such as confusing a pedestrian with a stationary object, can lead to engaging in a maneuver inappropriate for the situation with potentially catastrophic consequences. 

The presence of novel or unfamiliar objects can also lead to unpredictable and dangerous behaviors. This may be due to the inability of computer vision models to generalize as well as humans do. Human drivers can successfully tackle road conditions they have not encountered before but AI models are far less capable. This remains one of the open research challenges for the field of AI. The only known approach for mitigating the potential negative consequences from this gap in perception abilities is to train the AI models on large amounts of data representing every possible road condition. This problem is in principle very hard to solve once and for all due to many reasons, including high time and costs associated with collecting real data and labeling it, as well as the inherent difficulty of achieving complete coverage of all possible data scenarios \cite{song2023synthetic}.  Alternatively, the industry \cite{Nvidia_physicalAI_url} and researchers \cite{Cao2024,Yang2023ReconstructingOI} are looking for ways to generate synthetic data to use in training AI models for autonomous driving. However, this is another hard problem which can lead to worse performance \cite{Bai2024} in real life. 

Errors \cite{fatality_url, Tesla_pileup_url, tesla_failure_url, Tesla_fire_truck} similar to the ones mentioned above have already manifested in real-world accidents, underscoring the urgency of improving the robustness of AV perception. Such examples reveal the prevalence of object misdetection and misclassification that contribute to accidents, raising serious challenges to the acceptance of current AV technology. Although the field performance of self-driving cars approaches human driver's performance in some pilot deployments, the progress is uneven with respect to different system vendors and even the best ones are showing signs of brittleness when facing unusual road conditions~\cite{Cummins2024}. 

Our research aims to develop an assessment criteria for the robustness of object detection and classification used in autonomous driving perception systems. Our specific interest is in adversarial driving scenarios such as real-world object occlusions, intentional acts of vandalism,  as well as naturally occurring phenomena like varying weather and time of day. Weather impacts  both the driving environment and sensor data quality by impairing vehicle sensors. Our results suggest that factors such as weather, occlusions, vandalized traffic signs and time of day critically impact the quality and consistency of the model's predictions, which is quantified here using a deep ensemble method to measure multiple state-of-the-art models' agreement. Our results are primarily drawn from experiments using synthetic data in simulation, offering a fast and controlled environment. Simulation data alone is insufficient for real-world training in high-risk environments \cite{Bai2024}. Thus, our findings are qualitatively validated using real-world data collected by a test vehicle.

\subsection{Problem statement}

This research investigates the critical gap between the theoretical capabilities of AI perception systems and their real-world performance in AVs. Specifically, the following questions are explored:

\begin{itemize}
    \item How do varying environmental conditions, such as extreme weather, impact the robustness and accuracy of object detection and classification in AVs?
    \item How do naturally occurring or adversarially constructed occlusions/superpositions of objects on the road impact the performance of object detection and classification in AVs?
    \item Can synthetic data be leveraged to enhance the performance of autonomous vehicle perception systems in complex environments? 
\end{itemize}

These questions are addressed by developing a suite of CARLA \cite{carla} scenarios for assessing the predictive uncertainty under varied environmental conditions and adversarial conditions. A framework is proposed for processing real-time data, recorded data, and simulated camera data to assess the robustness of an ensemble of AI perception models. This framework enables the systematic evaluation of the perception system’s sensitivity to both environmental and adversarial parameters. Lastly, a set of tests for assessing the
 impact of object detection on
 vehicle control response, e.g., braking (Figure  \ref{fig:stoppingscenario}) are developed. 

The paper is structured as follows: Sec. \ref{sec:rel_work} provides a background on research efforts in uncertainty estimation for deep models and justifies the selection of deep ensembles for this research. Sec. \ref{subsec:safetyAssessment} introduces a simple stopping scenario and defines the criteria for perception system safety, including a perception assessment criterion and \textit{Safety Quadrant} ($\mathcal{R}$). Specifically, Sec. \ref{sec:uncertainty_quant_approach} details an ensemble-based approach that uses mean prediction probability ($\mu_{\mathcal{S}}$) and standard deviation ($\sigma_{\mathcal{S}}$) to heuristically estimate perception robustness. Sec. \ref{sec:methodology} outlines the experimental methodology, detailing the setup for both comprehensive simulation experiments (including baseline conditions, sensitivity analysis, and robustness limits evaluation) and supporting real-world testing. Sec. \ref{sec:results} presents the key findings of our experiments, highlighting the impact of varied naturally occurring and adversarial conditions on the robustness of object detection and classification. Sec. \ref{sec:discussion} discusses the implications of these findings for autonomous vehicle safety and Operational Design Domains (ODDs), analyzes the limitations of the study, and explores potential avenues for future research. Finally, Sec. \ref{sec:conclusion} concludes the paper by summarizing the key contributions and findings.

\section{Related Work and Background} \label{sec:rel_work}

Perception systems in AVs rely on data from multiple sensors such as Light Detection and Ranging (LiDAR), Radio Detection And Ranging (Radar), cameras, and ultrasonic sensors to create an understanding of its environment \cite{Silva2017FusionOL}. The combination of sensor data streams to leverage the strengths of each sensor modality is known as sensor fusion \cite{kocic2018}. Despite sensor fusion, inconsistent performance of perception systems remains a critical challenge, particularly under adverse conditions such as heavy rain, fog, snow or low-light scenarios \cite{zhang2023perception}. These conditions degrade sensor performance by introducing additional noise in the perception pipeline and reducing perception robustness \cite{sensible, Vargas2021, YONEDA2021}. For instance, LiDAR sensors may struggle to detect objects through heavy rain due to water droplet interference, while cameras may suffer from overexposure in sun glare or low visibility in fog. Moreover, high-risk environments, such as dense traffic in urban areas, altered traffic signs due to vandalism, or unstructured roads in rural settings, exacerbate these challenges.


To mitigate problems related to inconsistent performance, significant efforts have been made to develop uncertainty estimation methods that are compatible with deep learning models. These uncertainty estimation methods enable perception systems to generate confidence measures for their predictions, which can be crucial for downstream AV decision-making. The methods mainly fall under categories such as Bayesian, non-Bayesian, and Gaussian-based approaches. Bayesian approaches, like MC-Dropout \cite{grewal2024}, have proven useful in characterizing predictive distributions rather than providing point estimates, allowing for greater adaptability to outliers and extreme values \cite{bayesian}. However, such approaches are often computationally complex, especially when dealing with high-dimensional data or large models, hence making real-time deployment challenging. 


Non-Bayesian approaches such as Deep Ensembles \cite{deepEnsemblesNB} have shown promise in balancing effectiveness and efficiency \cite{deepEnsGoogle}, thus opening up the possibility of incorporating uncertainty estimation in AV perception systems as a means of enhancing safety and building fail-safe mechanisms \cite{grewal2024}. 
The diversity among ensemble members can reveal nuances that a single Gaussian distribution might miss. For instance, the ensemble approach might be able to capture multi-modal (more than one peak) as compared to a Gaussian approach. In addition to this, the ensembles are better at generalizing to new or unseen environments because they combine the strengths of multiple models with different features \cite{solovyev2021weighted, casado2020ensemble}. This often provides robustness under challenging conditions like rain, fog, spectral aberration, etc., thereby making class probability more confident and localization more precise.  Further, the computational cost, in terms of both the time as well as the resources is comparable between the ensemble and the Gaussian approaches.


Our work employs an ensemble of five heterogeneous models listed in Section~\ref{sec:CV models}. Although developing ensemble methods is not the focus of this work, we leverage the diversity among these models to analyze how overall variations in perception affect AV system performance.

\subsection{Computer Vision Models and Dataset} \label{sec:CV models}

Model selection is critical in machine learning (ML), especially when model outcomes directly influence time-critical, real-world decision-making.  Factors such as performance, real-time processing capabilities, computational complexity, robustness, and scalability must be carefully considered. Based on these criteria,  five state-of-the-art ML models for ensemble were selected, each offering unique strengths. Below is a brief overview of each model:

\begin{enumerate}
    \item \textbf{YOLO} \cite{Redmon_2016_CVPR} is a popular real-time object detection algorithm known for its high accuracy and speed. Unlike traditional methods, YOLO requires only a single pass for both object classification and regression tasks. This study utilizes two recent versions:
    
    \begin{enumerate}
        \item \textbf{YOLOv8} \cite{yolov8}: This state-of-the-art model offers enhanced performance and efficiency compared to previous YOLO versions.
        
        \item \textbf{YOLOv9} \cite{yolov9}: This version improves parameter and gradient utilization compared to YOLOv8, resulting in higher precision in benchmarks. (Note: The latest version, YOLOv10 \cite{wang2024yolov10}, was not included due to its relatively minor improvements over YOLOv9.)
    \end{enumerate}
    
    \item \textbf{DEtection TRansformer (DETR)} \cite{carion2020end}, introduced by Facebook AI Research, is based on a transformer encoder-decoder network. This model eliminates the need for non-maximum suppression (NMS). This work uses two DETR variants-DETR50 (trained with a ResNet50 backend) and DETR101 (trained with a ResNet101 backend).
    
    \item \textbf{Real-Time Detection Transformer (RT-DETR)} \cite{zhao2024detrs} is an end-to-end object detector that provides real-time performance with high accuracy. As a DETR \cite{carion2020end} derivative, it is suitable for applications requiring fast and accurate object detection, such as autonomous driving.
    \item \textbf{Gaussian YOLOv3 (YOLOv3)}\footnote{Since Gaussian YOLOv3 no longer reflects the performance compared to current object detection models, it has been excluded from the experimental comparisons.} \cite{choi2019gaussian} incorporates Gaussian parameters into YOLOv3. This model outputs localization uncertainty independent of the bounding box.
\end{enumerate}

All models were pre-trained on the MS COCO dataset \cite{lin2014microsoft} to ensure consistency in classes and facilitate a comparative analysis of their performance of object classification and detection when applied to the CARLA-generated dataset and real-world dataset. This pre-training approach provides a broad common foundation of the visual world, similar to the preliminary knowledge humans acquire before attending driving school, for evaluating the models' performance in diverse and challenging driving scenarios. Further, this road map provides an opportunity to assess the system's behavior as the distribution of roadway changes, thereby decoupling the model performance variation from training-induced bias.

\subsection{Driving Scenario for Perception Assessment}
\label{subsec:safetyAssessment}

\begin{figure}[H]
    \centering
    \includegraphics[width=0.95\linewidth]{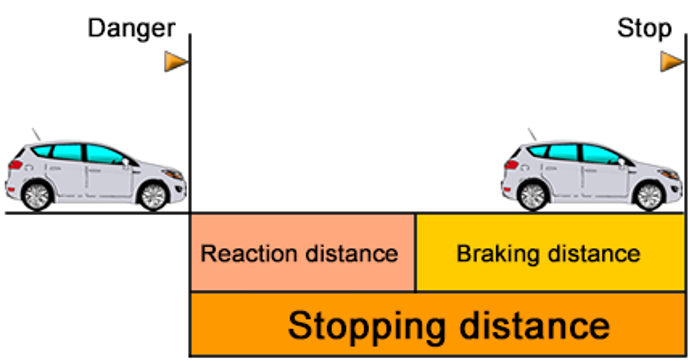}
    \caption{Simple Stopping Scenario}
    \label{fig:stoppingscenario}
\end{figure}

Consider a simple stopping scenario (Fig. \ref{fig:stoppingscenario}) where the vehicle must stop before reaching a stop sign on a straight path. Despite being a relatively straightforward traffic element, real-time detections of such signs are vital especially in dynamic or temporary traffic control situations such as near construction zones or near school buses where pop-up stop signs appear unexpectedly. Moreover, the high-definition (HD) maps that are pre-loaded in vehicle memory may not update instantaneously leading to environment discrepancies between the AV’s internal representation and the physical world. There might also be localization drift where AV's estimated position is slightly off therefore leading to missed detections or false positives. These challenges highlight the importance of robust perception-based stop sign detection and classification making this scenario both pragmatic and high-impact for autonomous vehicle perception evaluation.

A safe stop for a vehicle depends on two main factors: reaction distance ($rd$) and braking distance ($bd$). Together, they determine the stopping distance ($sd$). Equation~\ref{eq1}\footnote{Adopted from \url{https://tinyurl.com/44tadjsm} Note: The calculation done is for what's considered a \textit{``comfortable stop"}} demonstrates how these parameters are used to compute a safe stopping distance for the simple stopping scenario. Here, $s$ denotes the speed of the vehicle in kilometers per hour (KMPH), $r$ denotes the reaction time in seconds, and $f$ denotes the friction coefficient.

\begin{equation} \label{eq1}
\begin{split}
 rd & = \frac{s\times r}{3.6} \\
 bd & = \frac{s^2}{250 \times f} \\
sd & = rd + bd
\end{split}
\end{equation}

Typical reaction times ($r$) range from $0.5 - 2$ seconds for human drivers \cite{stoppingDistance} and from $0.82 - 0.93$ seconds for AVs on average with standard deviation of $0.54$ seconds in case of Google's and Mercedes-Benz AVs in California \cite{reactionTimes_dixit}. These reaction times for AVs can go as high as 8 seconds due to system disengagement \cite{IEEE_reactiontime_systemFailure} or vary depending on available compute resources on the platform and the algorithms used. The friction coefficient ($f$) is in the range $0.70 - 0.80$ for dry asphalt and  $0.1 - 0.6$ for wet or icy asphalt \cite{onspot2024}. 

For a human driver with reaction time $r \leftarrow 1$ sec and vehicle speed $s \leftarrow 48.29$ KMPH ($30$ miles per hour (MPH)), the equations yield,
$sd = 25.84$ meters on Dry Asphalt with $f \leftarrow 0.75$. In contrast, $sd = 50.72$ meters on Wet Asphalt with $f \leftarrow 0.25$ under assumption that vehicle tires and brakes are in good condition.

This example illustrates how a change in weather conditions can significantly affect the vehicle stopping distance. Moreover, the reaction time for the vehicle's perception module is determined by the amount of time the module will take to perform the tasks of object detection and classification.  Introducing additional factors, such as adversarial interventions, can further impact this distance by increasing the reaction time.    

\subsection{Model Disagreement and Sensitivity Estimation} \label{sec:uncertainty_quant_approach}


\begin{algorithm}[!ht]
\caption{Ensemble-Method}
\label{alg:uncertainty}
\begin{algorithmic}[1]
    \Require $(M, \mathcal{D}, \mathcal{S}, \check{\theta}_{\mathcal{S}} )$, where each model in $M$ is trained on dataset $\mathcal{D}$ \cite{lin2014microsoft}, $\mathcal{S}$ is the object class (stop sign) and $\check{\theta}_{\mathcal{S}}$ is the detection confidence threshold for $\mathcal{S}$ for participating models.
    \Ensure $|M| \geq 2$
    \State $\hat{M} \leftarrow \phi$ \Comment{Initialize ensemble}
    \State $\mu_{\mathcal{S}} \leftarrow \phi$, $\sigma_{\mathcal{S}} \leftarrow \phi$ \Comment{Initialize mean and standard deviation of the ensemble}
    \For{$\check{d} \in \mathcal{D}$}
        \For{$m$ $\in$ $M$}
            \If{$Pr_{\mathcal{S}}[\check{d}][m] \geq \check{\theta}_{\mathcal{S}}$} \Comment{$Pr_{\mathcal{S}}[\check{d}][m]$ is the prediction probability by $m$ on $\mathcal{S}$ for $\check{d}$}
                \State $\hat{M}[\check{d}][m] \leftarrow Pr_{\mathcal{S}}[\check{d}][m]$
            \Else 
                \State $\hat{M}[\check{d}][m] \leftarrow 0$
            \EndIf
        \EndFor
        \State $\mu_{\mathcal{S}}[\check{d}] \leftarrow$ mean for $\check{d}$ \Comment{$\forall m \in M$}
        \State $\sigma_{\mathcal{S}}[\check{d}] \leftarrow$  standard deviation for $\check{d}$ \Comment{$\forall m \in M$}
    \EndFor
    \State \textbf{Return} $(\hat{M}, \mu_{\mathcal{S}}, \sigma_{\mathcal{S}})$
\end{algorithmic}
\end{algorithm}

We use the ensemble approach presented in Algorithm~\ref{alg:uncertainty}, with  the intention of reducing the likelihood of errors due to overfitting or bias in individual models. Further, uncertainty for deep learning models is often difficult to estimate without incorporating specific design elements in the model's architecture, such as dropout layers or spectral normalization. Instead, the standard deviation of prediction probabilities of the ensemble can be used as a heuristic surrogate for the unknown uncertainties of the ensemble's constituent models.

In our experiments, we are interested in profiling the performance of the models and ensemble approach for a single class, i.e., a stop sign, denoted by $\mathcal{S}$. Let $M$ denote the set containing the models that participate in the ensemble approach, where each participating model is denoted by $m$, such that each  $m \in M$ is separately trained on dataset  $\mathcal{D}$ \cite{lin2014microsoft} and $\check{d} \in \mathcal{D}$ is a data sample. The cardinality of $M$, i.e., $|M|$, denotes the number of participating models. $\hat{M}$ denotes the ensemble of the models. Let $\check{\theta}_\mathcal{S}$ denote the \textit{detection confidence threshold} value of the class  $\mathcal{S}$ for participating model. If the participating model's prediction probability for  $\mathcal{S}$ is less than $\check{\theta}_\mathcal{S}$ $(=0.20)$, then the prediction is discarded by setting the probability to zero. $\hat{M}$ stores the prediction probability for each data sample $\check{d} \in \mathcal{D}$, and for each participating model $m \in M$. $\mu_{\mathcal{S}}$ and $\sigma_{\mathcal{S}}$ denote the mean prediction probability and the standard deviation of the ensemble for class $\mathcal{S}$, $\forall \check{d} \in \mathcal{D}$. 

Let $\theta_\mathcal{S}$ denote the threshold for the ensemble created using algorithm~\ref{alg:uncertainty}. Note, the threshold $\check{\theta}_\mathcal{S}$ for individual models and $\theta_\mathcal{S}$ for the mean class probability was selected based on specific conditions of our scenario and is not intended to serve as a universal safety benchmark but rather reflects maneuver-dependent and context-dependent nature of perception safety requirements. For e.g. a warehouse robot operating in an indoor structured environment may be able to function adequately with a lower class probability threshold, but an autonomous vehicle functioning at higher speeds would require significantly higher confidence threshold to ensure safety. 

Let $d_{\mathcal{S}}$ denote distance from object $\mathcal{S}$. If $sd \leq d_{\mathcal{S}}$ and $\mu_{\mathcal{S}} \geq \theta_{\mathcal{S}}$, where $sd \leftarrow (rd + bd)$, then the vehicle can comfortably stop before reaching the object $\mathcal{S}$. In contrast,  $\mu_{\mathcal{S}} < \theta_{\mathcal{S}}$, at any point when $sd \geq d_{\mathcal{S}}$, results in unsafe driving conditions that may be dangerous.

\subsection{Safety Quadrant} \label{sec:safety_quadrant}

Using Equation \ref{eq1}, the concept of a \textit{safety quadrant} (or region) is defined for the simple stopping scenario, denoted by $\mathcal{R}$. Consider a plot of $\mu_\mathcal{S}$ over $d_\mathcal{S}$. $\mathcal{R}$ is defined as the region where $\mu_{\mathcal{S}} \ge \theta_{\mathcal{S}}$ and $d_{\mathcal{S}} > sd$. This corresponds to the top-left quadrant of the plot when bisecting the plot with the intersecting lines $x=sd$ and $y=\theta_{\mathcal{S}}$. This quadrant signifies the ideal operating state where the perception system has achieved the required stable confidence while the vehicle is still at a distance greater than the stopping distance, allowing for a comfortable and safe stop.

For continued safe operation, once the system enters region $\mathcal{R}$, it is imperative that the trend of the mean prediction probability $\mu_{\mathcal{S}}$ should be non-decreasing, or at a minimum, remains above the threshold $\theta_{\mathcal{S}}$ as the distance $d_{\mathcal{S}}$ continues to decrease. Mathematically, for any time $t_1$ when $d_{\mathcal{S}}(t_1) \ge sd \land (\mu_{\mathcal{S}}(t_1) \ge \theta_{\mathcal{S}})$, and for any subsequent time $t_2 > t_1$ where $d_{\mathcal{S}}(t_2) < d_{\mathcal{S}}(t_1)$, it is required that $\mu_{\mathcal{S}}(t_2) \ge \mu_{\mathcal{S}}(t_1)$ or, at a minimum, $\mu_{\mathcal{S}}(t_2) \ge \theta_{\mathcal{S}}$. This is a critical temporal aspect for robust operation. Although variability and  perception uncertainty exist in the calculation of stopping distance and the establishment of the safety quadrant, for the purposes of assessing the sensitivity of the AV’s perception, these parameters are treated as fixed, thereby defining a canonical instance of the experiment.

A failure to satisfy this condition -- specifically, either failing to achieve $\mu_{\mathcal{S}} \ge \theta_{\mathcal{S}}$ when $d_{\mathcal{S}} > sd$ (i.e., mean prediction probability not entering region $\mathcal{R}$ by the stopping distance) or a significant drop in $\mu_{\mathcal{S}}$ value below the threshold $\theta_{\mathcal{S}}$ after initial entry into $\mathcal{R}$ -- raises a critical safety concern. Such a scenario could indicate inconsistencies or instabilities in the perception system's output, potentially leading to non-deterministic and potentially hazardous behaviors in downstream decision-making processes. For instance, a sudden decrease in the perceived probability of a critical object might cause the decision engine to prematurely disengage a safety maneuver (e.g., braking) based on an unreliable perception assessment of the environment. Therefore, ensuring the temporal consistency and persistence of high probability estimates within region $\mathcal{R}$ is crucial for the overall safe and reliable operation.

While robust perception performance in $\mathcal{R}$ is a necessary condition for the perception system, it is only one of several interconnected systems within the vehicle that determine the safe outcome in any driving scenario. For example, path planning and decision-making components of autonomous systems operate downstream of perception, relying on its output to execute safe driving maneuvers.

$\theta_{\mathcal{S}}$ is set to $0.75$ and $sd = 25.55$ m with respect to dry road conditions. 
Setting the ensemble threshold, i.e., $\theta_{\mathcal{S}}$, at 0.75, prioritizes detections with strong consensus across all participating models. This significantly reduces the false positive rate, which is critical in real-time driving scenarios where false alarms can degrade system performance or trigger erratic vehicle behavior. At the same time, moderately confident predictions from individual models are still able to contribute to $\mu_{\mathcal{S}}$, thereby limiting the risk of false negatives. The choice of $\theta_{\mathcal{S}} = 0.75$ also aligns with established practices in the research community, where confidence thresholds in the range of $0.70$ – $0.80$ are commonly adopted to filter low probability predictions in safety critical tasks such as autonomous driving~\cite{zhang2025auxdepthnet, yang2023mix}.
The distance $sd$ changes in real-world experiments based on speed. For example, $sd=5.84$ m at $10$ MPH increases to $sd=19.8$ m for $25$ MPH when coefficient of friction  is equal to $0.75$.




\section{Methodology}\label{sec:methodology}

\begin{figure*}[!ht]
    \centering
    \includegraphics[scale=0.7]{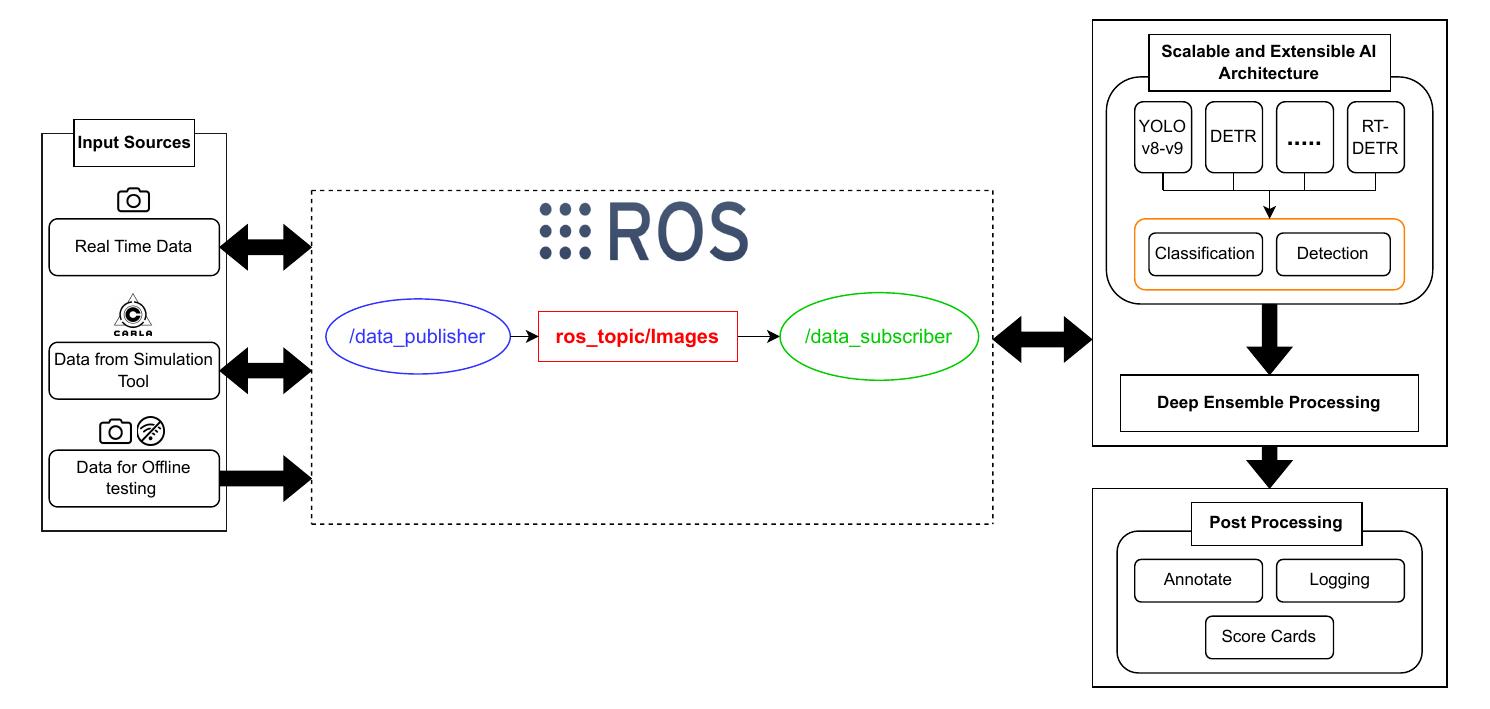}
    \caption{\textit{Experiment Architecture:} Data from input sources (left) -- real-time, CARLA, or offline rosbags -- communicated to scalable and extensible AI (right) through ROS framework (center). The AI models perform object classification and detection, deep ensemble processing and additional post processing (right) and communicate  back the results.}
    \label{fig:arch}
\end{figure*}


As outlined in the research objectives, a central goal of this work is to establish reliable measurement criteria for the robustness of perception systems in autonomous vehicles facing adversarial scenarios. To achieve this, the experimental setup incorporates the generation of synthetic data specifically designed to model real-world object occlusions. Crucially, this research emphasizes the necessity of calibrating synthetic data with real-world experiments to avoid the pitfalls associated with exclusive reliance on simulated data for training and evaluation in safety-critical applications.

\subsection{Experiment Architecture} \label{sec:setup experiment}

Figure \ref{fig:arch} depicts the overall experiment architecture\footnote{Source Code: \href{https://github.com/usnistgov/NIST_AV_AI}{https://github.com/usnistgov/NIST\_AV\_AI}} used in this work. This architecture integrates data from different input sources, including real-time vehicle sensor data, CARLA-generated synthetic data, and offline datasets (e.g. ROS bags), allowing for comprehensive benchmarking, controlled experimentation relevant to adversarial conditions, and real-world validation through a single interface. The architecture features an underlying abstraction layer, called the robot operating system (ROS), which facilitates communication with perception models, enabling seamless interaction across diverse data streams for scenario-based testing. For instance, real-time data from vehicle sensors can be used to test the system's responsiveness in real-world driving conditions, while offline datasets (\textit{rosbags}) allow for efficient testing and analysis of pre-recorded scenarios. 

Specifically, CARLA \cite{carla} provides the capability to generate synthetic data, introducing both natural driving conditions (e.g., heavy rain, fog, nighttime) and adversarial conditions (e.g., occluded, altered or vandalized traffic signs). Data generated within CARLA, processed via ROS middleware, is then fed into our ensemble of object detection models. This enables comprehensive testing of the AI models under diverse and challenging circumstances. 


For real-world experiments, data was collected using a Lucid camera\footnote{Camera model Lucid TRIO16S-CC: \href{https://tinyurl.com/lucid-camera-specs}{https://tinyurl.com/lucid-camera-specs}}. To ensure consistency with our simulation environment, we closely matched simulation parameters to the real-world conditions under which the Lucid camera data was captured. Vehicle sensor placement in  CARLA follows the KITTI data collection platform \cite{KITTI}, and the sensor configurations are designed to closely match those of the real vehicle. 

While our current experiments focus on detecting a stop sign, the developed system can be adapted for the detection of other dynamic objects or different objects of interest on the road. 

\subsection{The Base Experiment Scenario in Simulation}

\begin{figure}[!ht]
    \centering
    \includegraphics[width=.45\textwidth]{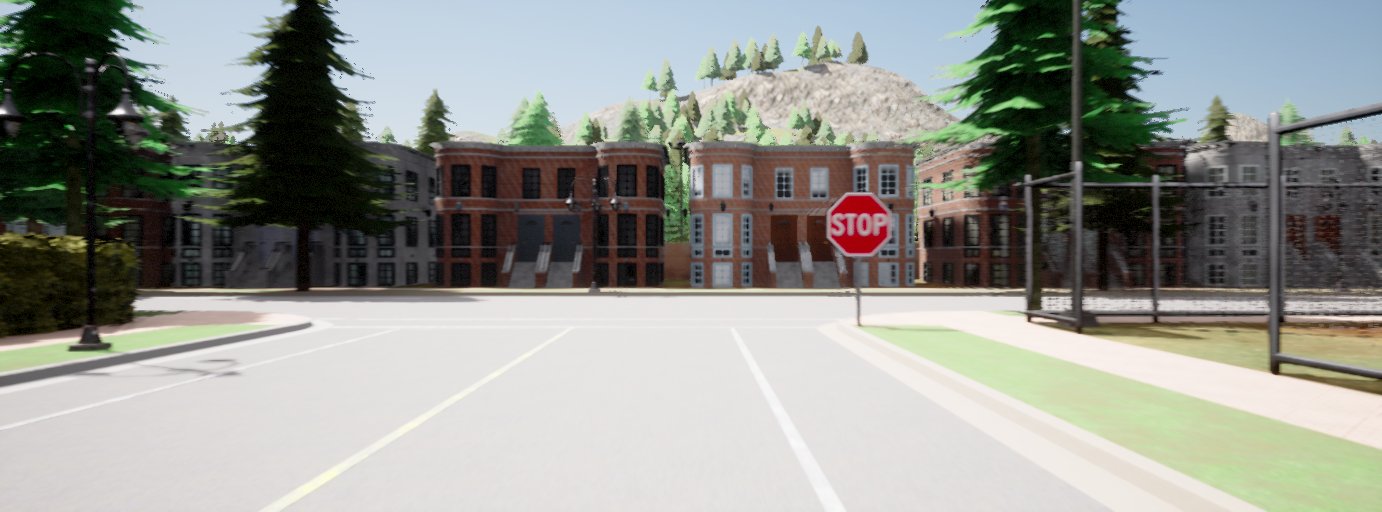}
    \caption{\textit{Base Experiment:} View from the ego vehicle (\textit{ego}) approaching the stop sign $\mathcal{S}$ in CARLA}
    \label{fig:base_exp}
\end{figure}

The base experiment scenario, which serves as the control condition, features a single Ego Vehicle  (\textit{ego}) instructed to move forward at a constant speed of 30 MPH along a central road, approaching a stop sign $\mathcal{S}$. Figure \ref{fig:base_exp} illustrates an instance of this baseline experiment, showing the front camera view from the ego in CARLA when the vehicle is 16.55 m away from $\mathcal{S}$.

\begin{table}[ht]
    \centering
    \caption{Weather parameters used in the first experiment set.}
    \label{tab:exp1:WeatherParam}
    \begin{tabularx}{\linewidth}{p{1.5cm} l X}
        \hline
        \textbf{Name} & \textbf{Values} & \textbf{Description}\\
        \hline
        Cloudiness & $0\%, 33.33\%, 66.67\%$ & Simulated clouds, which reduce brightness \\
        Precipitation & $0\%, 33.33\%, 66.67\%$ & Simulated puddles, rain particles, and water droplets on cameras \\
        Fog Density & $0\%, 33.33\%, 66.67\%$ & Simulated fog, which reduces visible distance \\
        Sun Azimuth Angle & $0^\circ, 120^\circ, 240^\circ$ & Moves sun position laterally (impacts glare, reflections, shadow direction etc.) \\
        Sun Altitude Angle & $-10^\circ, 23.33^\circ, 56.67^\circ$ & Moves sun position vertically (primarily impacts time of day, shadow length and overall brightness) \\
        \hline
    \end{tabularx}
\end{table}

\begin{table}[ht]
    \centering
    \caption{Adversarial Parameters Used in the Simulation}
    \label{tab:AdverseParam}
    \begin{tabularx}{\linewidth}{>{\raggedright\arraybackslash}p{2cm} X}

        \hline
        \textbf{Name} & \textbf{Description} \\
        \hline
        Ambulance 1 & An ambulance positioned to cause low occlusion of the stop sign. \\
        Ambulance 2 & An ambulance positioned to cause moderate occlusion of the stop sign. \\
        Firetruck & A firetruck positioned behind the stop sign, causing perceptional camouflage and reverse visibility issues. \\
        Graffiti~\cite{pavlitska2023adversarial} & Black spray paint tactically applied to obscure the stop sign's visibility. \\
        Tree & A tree strategically placed to occlude the stop sign. \\
        Pole & A pole partially obstructing the stop sign. \\
        Pedestrian & A pedestrian partially blocking the stop sign from view. \\
        Adversarial Patch \cite{pavlitska2023adversarial, tsuruoka2024wip} & A texture patch applied to the stop sign to degrade perception network performance. \\
        Low-Res Patch \cite{pavlitska2023adversarial, tsuruoka2024wip} & A down-sampled version of the adversarial patch. \\
        Bush 1 & A bush placed in front of the stop sign to induce occlusion. \\
        Bush 2 & An alternative bush model used for stop sign occlusion. \\
        Alt. Stop & An alternative low resolution stop sign model used in the simulation. \\
        \hline
    \end{tabularx}
\end{table}


\begin{figure*}[ht!]
    \centering
    \subfloat[No Occlusion: Stop sign ($\mathcal{S}$) fully visible.]{\includegraphics[width=0.48\linewidth]{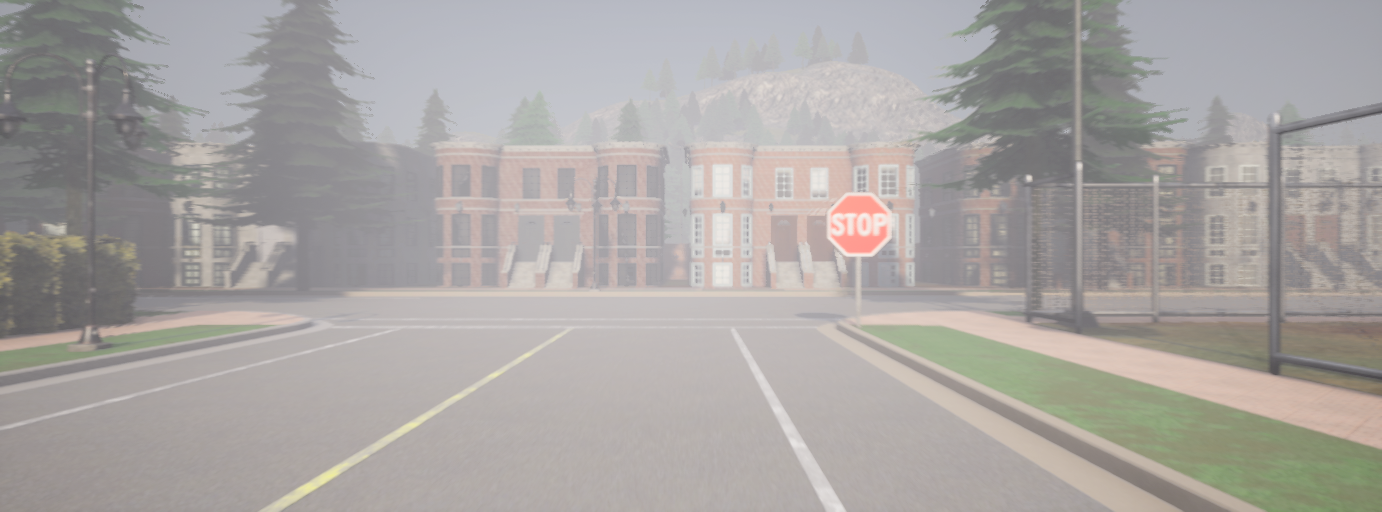}\label{fig:occlusion_no}}
    \hfill 
    \subfloat[Low Occlusion: $\mathcal{S}$ minimally obscured by an adversarial object.]{\includegraphics[width=0.48\linewidth]{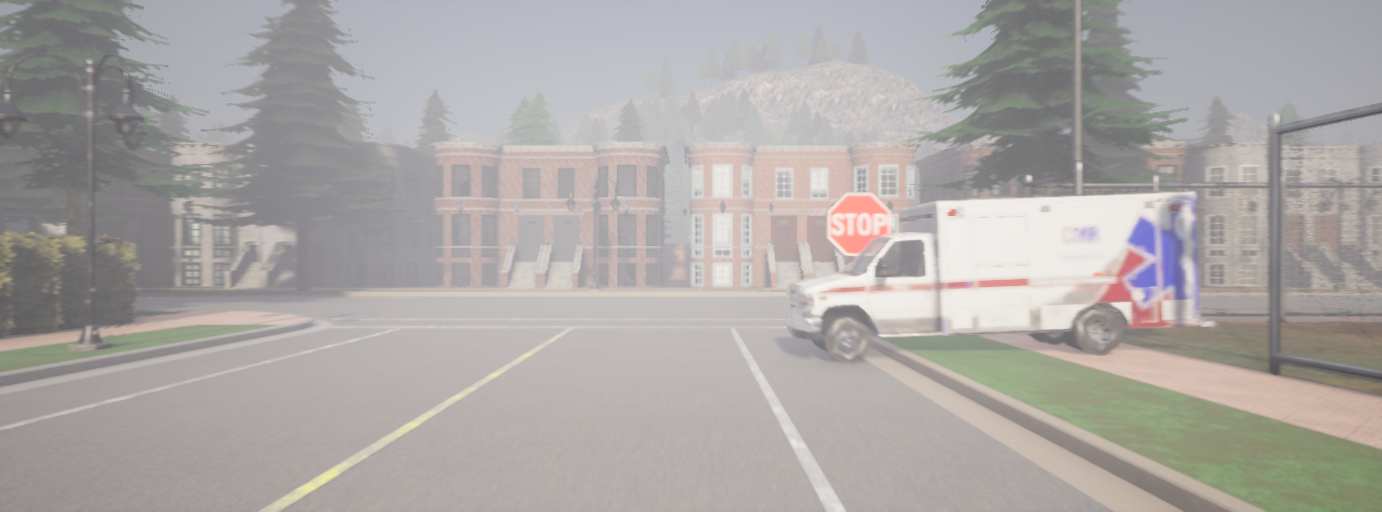}\label{fig:occlusion_low}}

    \vspace{0.5em} 

    \subfloat[Moderate Occlusion: $\mathcal{S}$ partially obscured by an adversarial object.]{\includegraphics[width=0.48\linewidth]{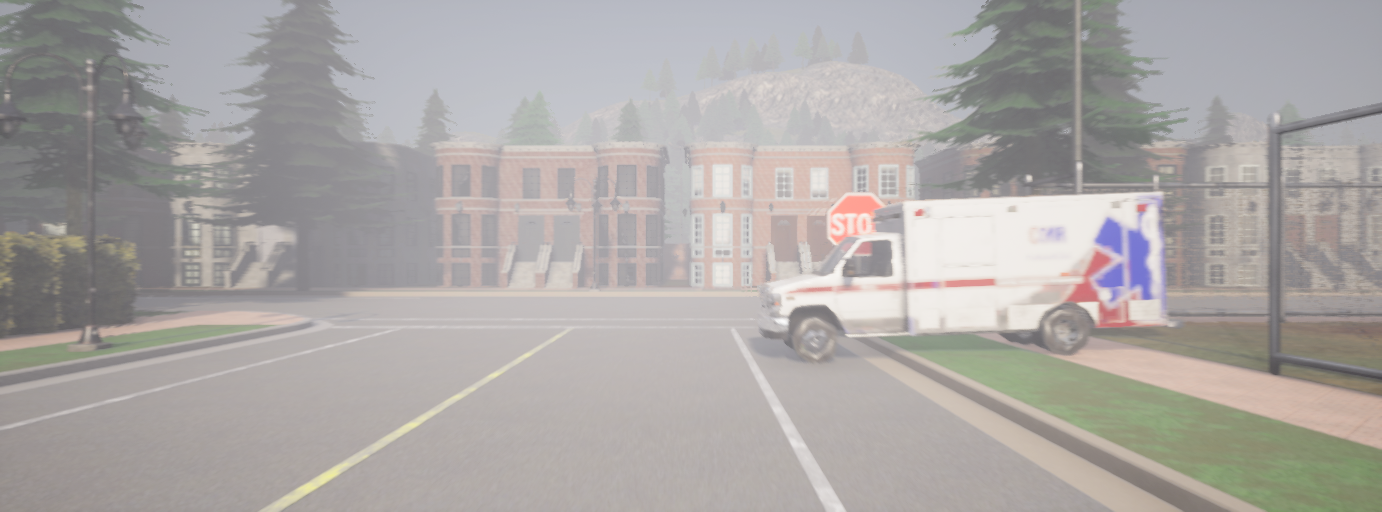}\label{fig:occlusion_mod}}
    \hfill 
    \subfloat[High perceptual camouflage: $\mathcal{S}$ significantly obscured by an adversarial object.]{\includegraphics[width=0.48\linewidth]{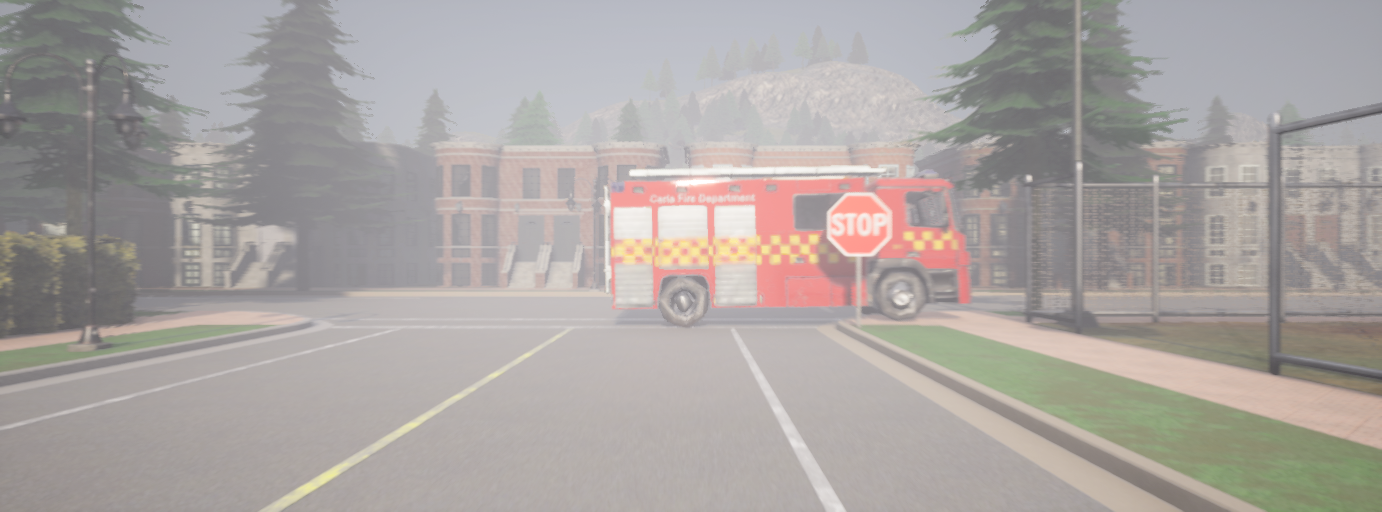}\label{fig:occlusion_high}}

    \caption{Representative CARLA simulation scenes demonstrating the four distinct occlusion levels used in Experiment Set 1, as viewed from the ego. Each image captures the ego's perspective approaching the stop sign ($\mathcal{S}$) at the same distance and under identical weather conditions, showcasing varying degrees of obstruction by adversarial objects.}
    \label{fig:occlusion_levels}
\end{figure*}

Key state variables were systematically manipulated in subsequent experiments. Each change in a state variable constituted a distinct experiment to allow individual analysis of its effect. These state variables encompassed the weather parameters (listed in Table \ref{tab:exp1:WeatherParam}) and the presence and characteristics of various adversarial objects and actors introduced to the simulated scene as occlusions (detailed in Table \ref{tab:AdverseParam}).  Following are the weather parameters for the base experiment scenario: cloudiness $= 0.0\%$, precipitation $= 0.0\%$, fog density $= 0.0\%$, sun azimuth angle $= 120^\circ$ and sun altitude angle $= 56.67^\circ$. The base experiment does not include any occlusions as seen in Figure \ref{fig:base_exp}.

\begin{figure*}[ht]
    \centering
    \subfloat[No Occlusion]{\includegraphics[width=0.48\linewidth]{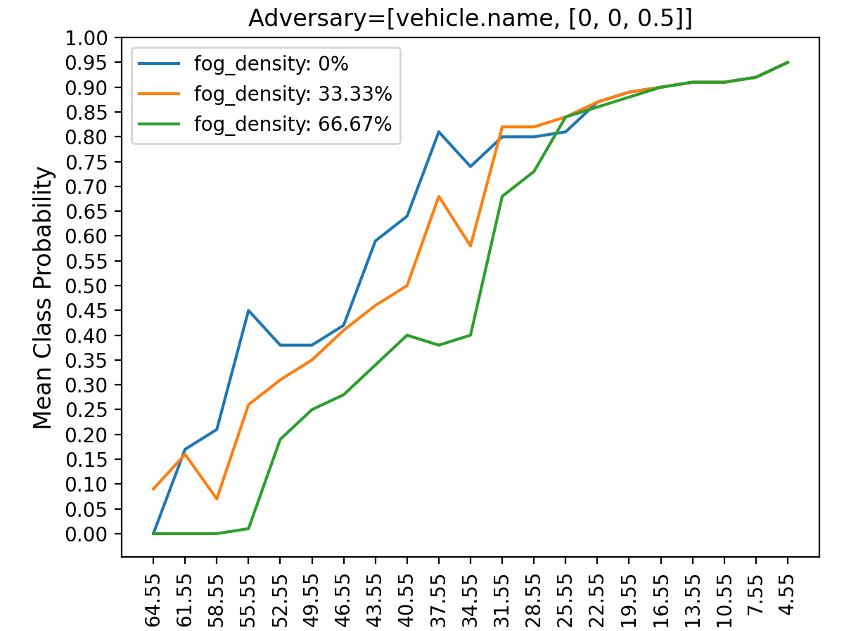}\label{fig:exp1:fog_noAdv}}
    \hfill 
    \subfloat[Low Occlusion]{\includegraphics[width=0.48\linewidth]{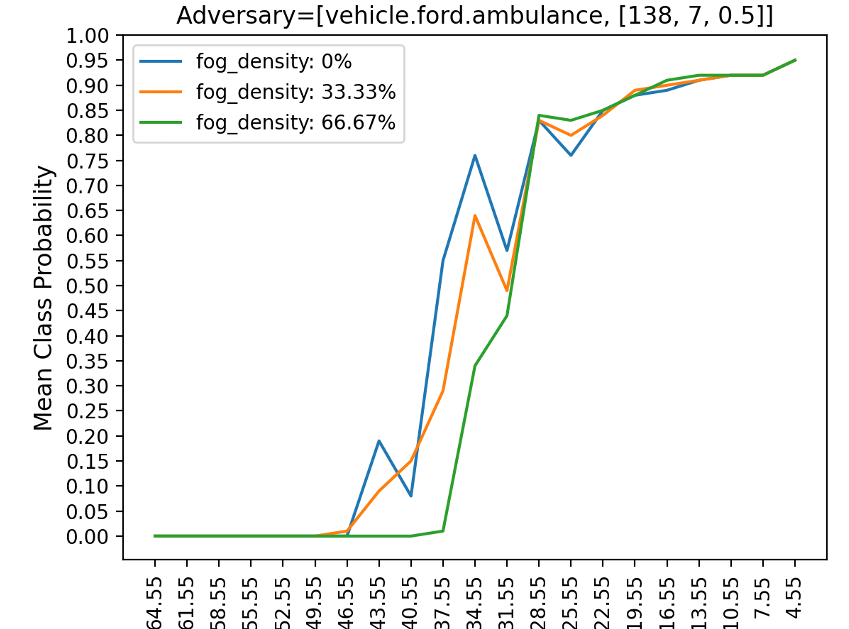}\label{fig:exp1:fog_lowAdv}}

    \vspace{0.5em} 

    \subfloat[Moderate Occlusion]{\includegraphics[width=0.48\linewidth]{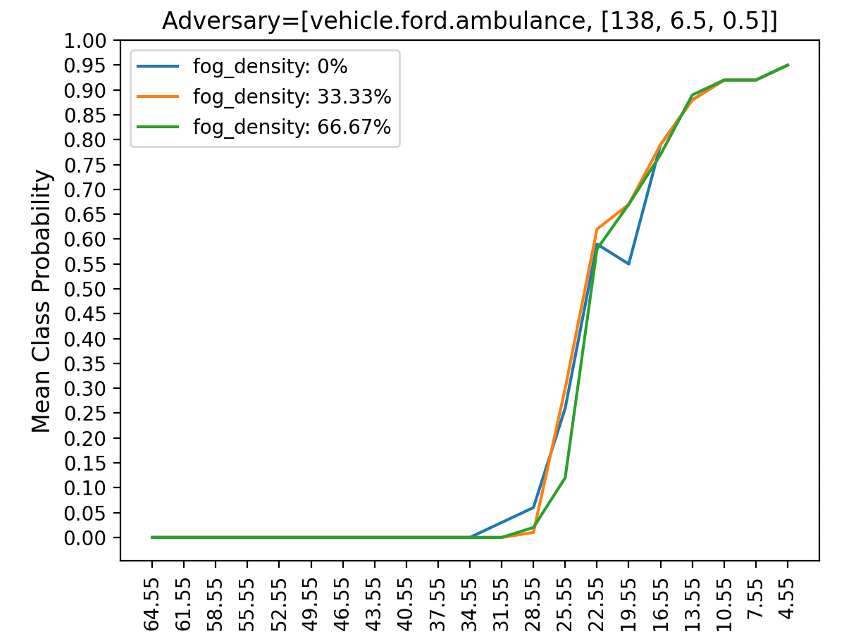}\label{fig:exp1:fog_modAdv}}
    \hfill 
    \subfloat[High Occlusion]{\includegraphics[width=0.48\linewidth]{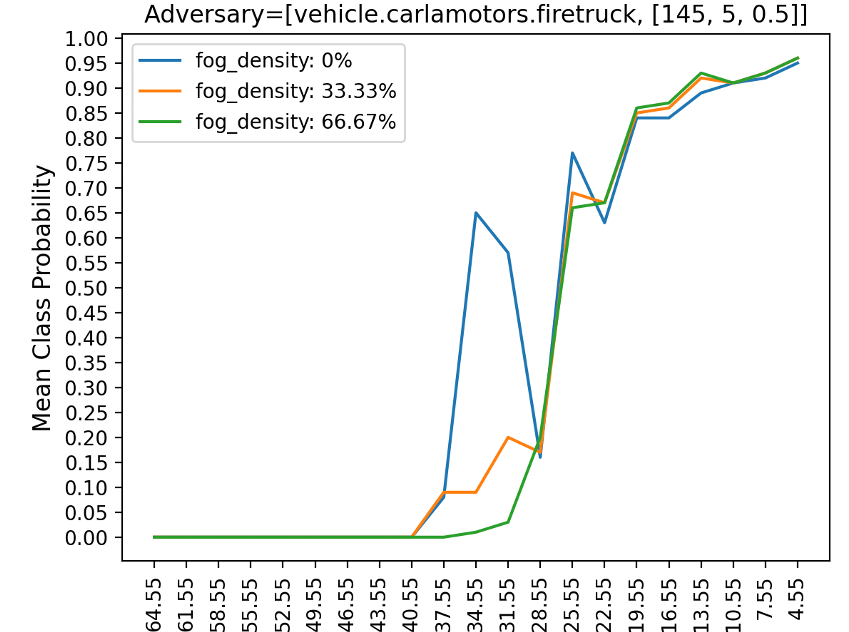}\label{fig:exp1:fog_highAdv}}

    \caption{Mean Prediction Probability ($\mu^{Y}_\mathcal{S}$) of YOLOv8 for Stop Sign ($\mathcal{S}$) Detection. Each subgraph displays $\mu^{Y}_\mathcal{S}$ (y-axis) as a function of distance from $\mathcal{S}$ (x-axis) under varying fog densities (0\%, 33\%, 66\%) for a specific occlusion level.}
    \label{fig:exp1-fog-density}
\end{figure*}

All experiments, including the base experiment and those with modified state variables, adhered to a pre-calculated stopping distance ($sd$) detailed in Sec. \ref{subsec:safetyAssessment} and maintained a detection confidence threshold of $\theta_{\mathcal{S}} = 0.75$, ensuring consistency across all evaluations. Furthermore, the ego's speed, driving behavior, roadway, and driving direction remained unchanged from the base scenario. This controlled approach ensured that any observed changes in the perception system's predictions could be directly attributed to the specific modifications in the state variables, relative to the established control.

The experimental evaluation was broadly divided into two distinct sets, each with a specific objective. \textbf{Experiment Set 1} aimed to identify and isolate specific weather parameters that most influenced the detection and localization task under three levels of occlusion (low, moderate, and high), introducing three specific adversarial objects. Building on these findings, \textbf{Experiment Set 2} uses a more refined sweep of weather parameters based on their large impact on prediction standard deviation, as well as a larger set of adversarial parameters.

\subsection{Sensitivity Analysis of Perception System Robustness} \label{sec:experiment1}

This first set of experiments focused on identifying the key factors contributing to high prediction standard deviation in the perception ensemble. It systematically investigated the impact of varied weather parameters (Table \ref{tab:exp1:WeatherParam}) and three specific adversarial objects (detailed in Table \ref{tab:AdverseParam}) designed to induce low, moderate, and high occlusions on the detection and localization of the stop sign ($\mathcal{S}$) within simulated driving scenarios. This exploration determined how these variations and occlusions impact the ego's perception system, consequently affecting its downstream decision-making for safe operation within the simple stopping scenario.

The primary objective of this experimental set was twofold:
\begin{itemize}
    \item \textbf{Conduct sensitivity and vulnerability analysis:} This involved identifying the specific weather and adversarial parameters that most significantly impacted the perception system. This sensitivity analysis aimed to pinpoint the system's vulnerabilities under varying conditions, thereby informing the design of challenging test scenarios for the second experiment set.
    
    \item \textbf{Evaluate model robustness:} This involved assessing the effect of state variables on the performance of the participating models and the ensemble. Robustness was approximated by the ensemble's prediction standard deviation using Algorithm \ref{alg:uncertainty}. Each experiment was assessed to determine if the ensemble operated within $\mathcal{R}$, specifically by verifying if the ensemble's ($\mu_{\mathcal{S}}$) reached or exceeded the defined confidence threshold ($\theta_{\mathcal{S}} = 0.75$) before the vehicle reached the safe stopping distance ($sd = 25.55$~m).
\end{itemize}

To achieve these objectives, each participating model (YOLOv8, YOLOv9, DETR50, DETR101, and RT-DETR) was systematically profiled across various weather parameters and adversarial conditions. This profiling generated detailed performance metrics for the perception system. Specifically, $\mu_{\mathcal{S}}$ and $\sigma_{\mathcal{S}}$ for the ensemble were analyzed to assess robustness under diverse environmental and adversarial conditions. This data then directly informed the perception assessment mentioned in Sec. \ref{subsec:safetyAssessment}, evaluating if the ensemble's collective confidence entered and consistently remained within $\mathcal{R}$ as the ego approached $\mathcal{S}$. For illustrative purposes, the dominant weather parameters are presented using YOLOv8 results, as similar trends were observed across other models.

To clarify the nature of the adversarial occlusions introduced in these experiments, Figure \ref{fig:occlusion_levels} visually presents representative examples from the CARLA simulation environment. This figure illustrates the four distinct occlusion scenarios investigated: no occlusion, low occlusion, moderate occlusion, and high occlusion, the latter three corresponding to specific adversarial object placements designed to progressively obscure the stop sign ($\mathcal{S}$). This visual context is crucial for understanding the subsequent analysis of model performance under these challenging conditions.

Figure \ref{fig:exp1-fog-density} presents a specific set of results from the sensitivity analysis, illustrating the impact of varied fog density on YOLOv8's mean prediction probability ($\mu^{Y}_\mathcal{S}$) across different occlusion levels. This figure comprises four graphs, each dedicated to one of the occlusion scenarios: no occlusion, low occlusion, moderate occlusion, and high occlusion, as visually defined in Figure \ref{fig:occlusion_levels}. Within each graph, three distinct trend lines depict $\mu^{Y}_\mathcal{S}$ (y-axis) at different distances from the stop sign ($\mathcal{S}$) (x-axis), corresponding to the three discrete fog density settings (0\%, 33\%, and 66\%) as detailed in Table \ref{tab:exp1:WeatherParam}. $\mu^{Y}_\mathcal{S}$ for each graph is computed by aggregating all scenarios corresponding to a given level of occlusion under the respective fog density settings.

This representative example demonstrates the systematic analysis conducted to understand the influence of specific environmental and adversarial conditions on object detection performance. The data reveals that as fog density increases from 0\% to 66.67\%, YOLOv8's performance consistently degrades across all four adversarial configurations. Furthermore, the presence of an adversary introduces a delay in stop sign classification, indicating a significant impact of occlusion on model performance.

Additionally, the effect of individual weather parameters is investigated through heatmaps. Each heatmap compares two parameters to illustrate their combined effect on ensemble standard deviation ($\sigma^{Y}_{\mathcal{S}}$). This generated a total of 10 heatmaps per model. Again, results are presented for YOLOv8. Profiling $\sigma^{Y}_{\mathcal{S}}$, rather than focusing solely on mean classification probability, provides some intuition as to the ensemble's sensitivity analysis, with higher $\sigma^{Y}_{\mathcal{S}}$ indicating performance degradation.

\begin{figure}[!ht]
    \centering
    \includegraphics[width=0.95\linewidth]{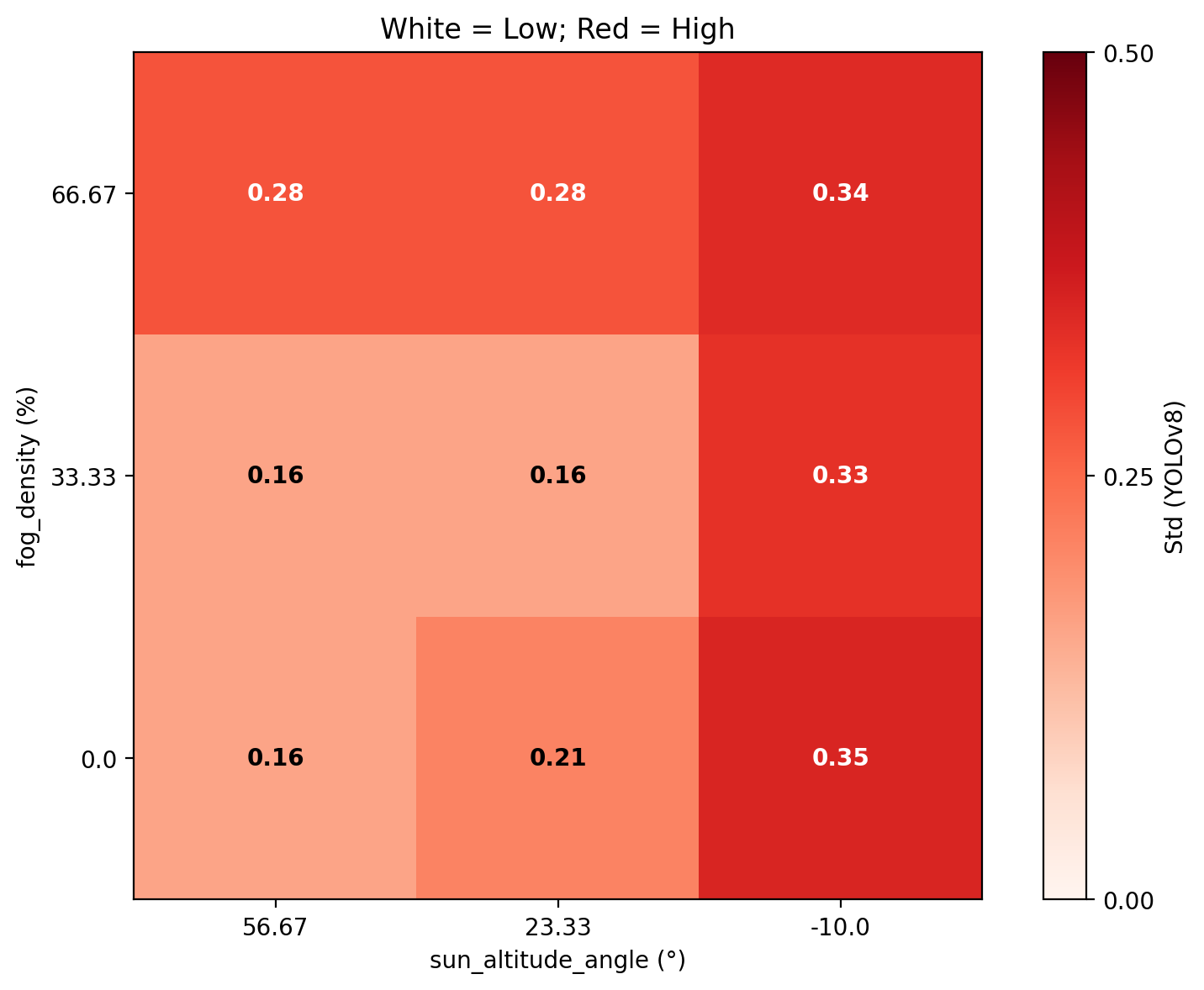}
    \caption{Heatmap illustrating ($\sigma^{Y}_{\mathcal{S}}$) for YOLOv8 as a function of Fog Density (y-axis) and Sun Altitude Angle (x-axis) in Experiment Set 1. Higher values (e.g., warmer colors) indicate worse ensemble performance.}
    \label{fig:heatmap:experiment1:f-sal}
\end{figure}

For example, Figure \ref{fig:heatmap:experiment1:f-sal} presents a heatmap showing the combined effect of fog density and sun altitude angle. This heatmap reveals that the YOLOv8 model exhibits increased $\sigma^{Y}_{\mathcal{S}}$ as fog density approaches its maximum value (66.67\%) and sun altitude angle reaches its lowest value ($-10^\circ$), corresponding to nighttime conditions. High cloudiness can similarly induce low lighting conditions, potentially yielding comparable effects on model performance.

The systematic analysis conducted in this first experimental set, exemplified by the impact of increasing fog density and varying sun altitude angles, successfully identified key state variables that significantly influence the perception system's detection and localization performance. These findings indicate that extreme weather conditions, particularly high fog densities and low light environments (e.g., nighttime), combined with higher levels of occlusion, significantly degrade object detection performance.

\subsection{Robustness Evaluation Under Intensified Environmental Conditions}

\textbf{Experiment Set 2} increased the gradation and overall severity of key weather parameters, while also incorporating additional adversarial occlusion cases. The primary objective was to demonstrate performance degradation under these severe conditions and to assess how such challenges degrade the overall robustness of the autonomous vehicle's perception capabilities.

To achieve these intensified operating conditions, the range of weather parameters was expanded to utilize the full capabilities of the CARLA simulation environment. Specifically, cloudiness, precipitation, and fog density were varied across their full 0 -- 100\% range, sun azimuth angle spanned 0 -- 360$^\circ$, and sun altitude angle covered -90 -- 90$^\circ$. This comprehensive exploration aimed to cover a wider spectrum of adverse environmental conditions known to challenge perception systems. 

Furthermore, complementing the three initial occlusion levels from the first set of experiments, the second set incorporated the remaining adversarial objects listed in Table \ref{tab:AdverseParam}. These included various naturally occurring obstructions like partial obscuration by poles, bushes, or trees, as well as altered signs featuring graffiti or patches, all derived from existing literature. This strategic selection of intensified parameters and complex adversarial scenarios was designed to systematically push the perception system's limits and induce extreme levels of predictive sensitivity.

During this experimentation phase, the perception system's performance was rigorously evaluated by applying the same ensemble-based framework detailed in Algorithm \ref{alg:uncertainty} and Section \ref{sec:uncertainty_quant_approach}. This comprehensive evaluation aimed to precisely identify the thresholds and combinations of extreme environmental and adversarial parameters that lead to quantifiable degradation in perception performance, thereby pushing the established boundaries necessary for safe operation within the region $\mathcal{R}$.

\subsection{Real-World Experimentation}
Given the importance of validating simulated findings with real-world observations, a limited set of experiments was conducted using a physical vehicle. While these real-world tests were not as extensive or comprehensive as the simulation-based analysis, primarily due to inherent physical and logistical constraints, they provided crucial data for calibrating the synthetic environment and preventing an exclusive reliance on simulated data for safety-critical evaluations. Importantly, these real-world observations also confirmed perception degradation, aligned with the expectations based on the simulated environments. This approach ensured that the insights gained from the comprehensive simulated scenarios maintained relevance to real-world operational conditions.         
\section{Results and Analysis}
\label{sec:results}

The experimental evaluation comprised a substantial number of unique scenarios, each representing a distinct configuration of environmental or adversarial state variables. A total of 972\footnote{A total of $972 \;(=3^5 \times 4)$ scenarios are obtained by varying cloudiness, rain, sun altitude angle, sun azimuth angle, and fog density, each at three discrete levels, combined with four types of adversarial conditions. See Tables~\ref{tab:exp1:WeatherParam} and \ref{tab:AdverseParam} for detailed parameter values.} such scenarios were conducted in Experiment Set 1, with an additional 1056\footnote{The scenarios in the Experiment Set 2 are obtained by adjusting the discretization of relevant parameters from Experiment Set 1 and by introducing additional adversaries.} scenarios in Experiment Set 2. Across these comprehensive tests, a significant portion exhibited robust perception performance. Characterized by the ensemble's mean prediction probability ($\mu_{\mathcal{S}}$) reliably exceeding the confidence threshold or borderline prediction ($\theta_{\mathcal{S}} = 0.75$, a value specifically selected for these experiments) before the stopping distance ($sd = 25.55$ m, calculated for the defined dry road conditions) and maintaining this confidence with a narrowing standard deviation ($\sigma_{\mathcal{S}}$) band, these results establish a foundational reference for normal system behavior. Understanding the perception system's expected behavior under non-challenging conditions is crucial for interpreting subsequent experimental results.

\subsection{Baseline Performance}
\label{sssec:results:baseline_performance}

\begin{figure}[!ht]
    \centering
    \includegraphics[width=\linewidth]{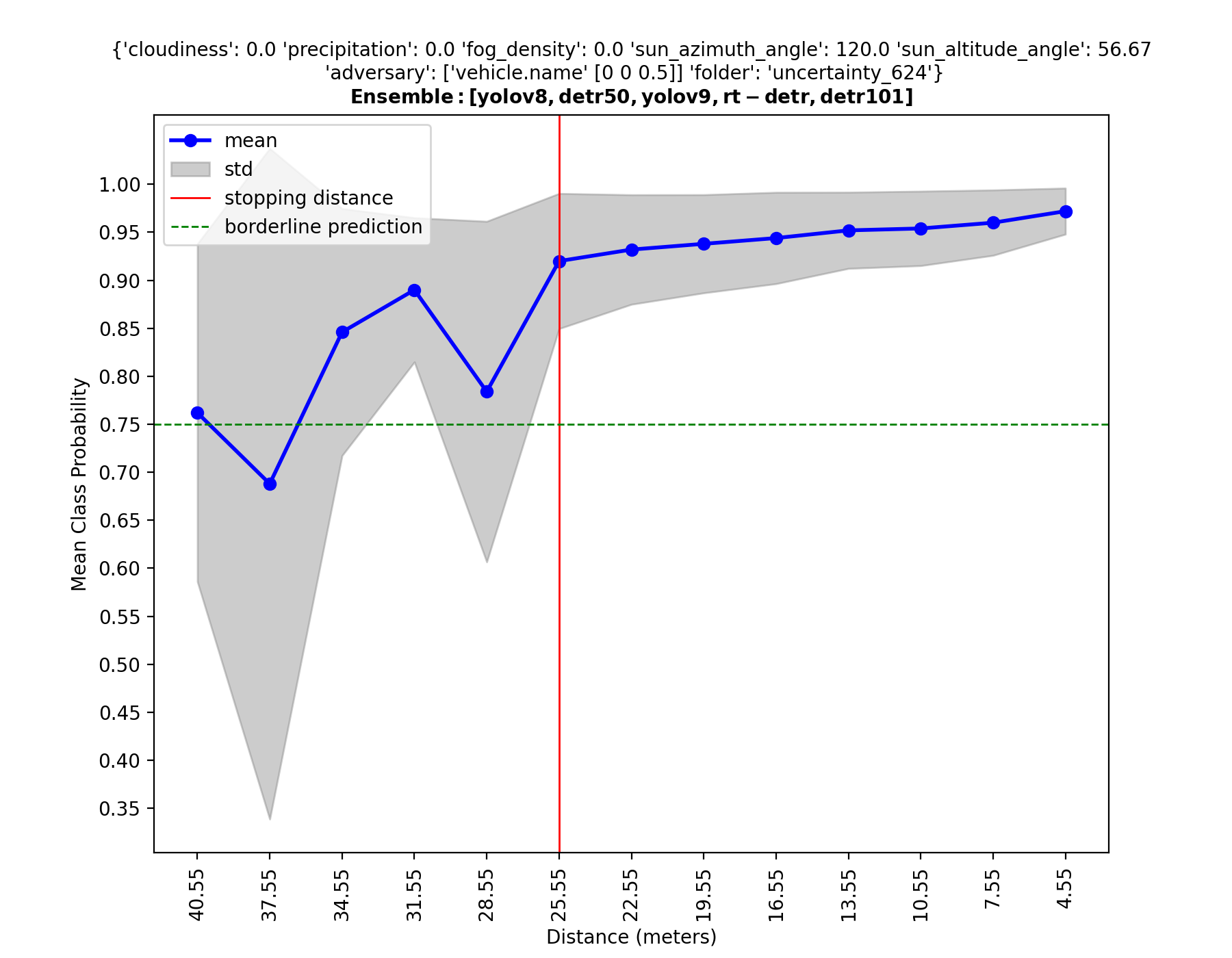} 
    \caption{Baseline Perception Performance: Mean prediction probability ($\mu_{\mathcal{S}}$) and standard deviation ($\sigma_{\mathcal{S}}$) for stop sign detection under optimal conditions (0\% cloudiness, 0\% precipitation, 0\% fog density, clear sun angle, no occlusion). The graph visually demarcates the safety quadrant $\mathcal{R}$, defined by the intersection of the vertical red line (stopping distance, $sd = 25.55$ m) and the horizontal green dotted line (confidence threshold, $\theta_{\mathcal{S}} = 0.75$). The gray shaded area represents $\sigma_{\mathcal{S}}$.}
    \label{fig:baseline_performance_graph}
\end{figure}

Figure \ref{fig:baseline_performance_graph} illustrates the baseline performance of the ensemble in a scenario characterized by optimal weather (0\% cloudiness, 0\% precipitation, 0\% fog density, and a clear sun angle) and no adversarial occlusions. This graph serves as a representative example of the many scenarios observed in Experiment Set 1 that exhibited robust and low predictive sensitivity behavior. This graph visually demarcates the safety quadrant $\mathcal{R}$, formed by the intersection of the $sd$ and $\theta_{\mathcal{S}}$ lines. 

As depicted, the mean prediction probability ($\mu_{\mathcal{S}}$) for the stop sign ($\mathcal{S}$) consistently remains high, reliably exceeding the confidence threshold ($\theta_{\mathcal{S}} = 0.75$) well before the safe stopping distance ($sd = 25.55$ m) mark. Furthermore, $\mu_{\mathcal{S}}$ exhibits a clear monotonically increasing trend as the ego approaches $\mathcal{S}$, and the surrounding standard deviation ($\sigma_{\mathcal{S}}$) band significantly narrows. This narrowing indicates decreasing predictive sensitivity and increasing confidence in the ensemble's prediction as the object becomes more clearly perceptible. This consistent performance serves as the reference point for assessing the degradation observed under more challenging conditions.

Other scenarios, however, revealed significant degradation under varied operating conditions. Sections \ref{subsec:results:exp1} and \ref{subsec:results:exp2} below provide representative case studies for these situations.

\subsection{Sensitivity Analysis: Evidence of Perception Variability}
\label{subsec:results:exp1}

As identified earlier in Sec. \ref{sec:experiment1}, fog density and sun altitude angle emerged as the most impactful weather parameters on object detection and localization performance. The following discussion presents the key observations from these experiments, illustrating how these varied conditions influence the ensemble's mean prediction probability and perception sensitivity.

While a significant number of scenarios within Experiment Set 1 demonstrated robust perception performance consistent with the established baseline (Section \ref{sssec:results:baseline_performance}), the sensitivity analysis revealed critical operating conditions that led to increased model disagreement. Key indicators observed include notable fluctuations in the mean prediction probability ($\mu_{\mathcal{S}}$), wider than normal standard deviation ($\sigma_{\mathcal{S}}$) bands -- particularly within region $\mathcal{R}$ -- and failures to consistently meet the confidence threshold ($\theta_{\mathcal{S}}$) after reaching stopping distance ($sd$).

The systematic investigation in Experiment Set 1 revealed a clear degradation in perception performance directly correlated with specific environmental and adversarial parameters. As illustrated in Figure \ref{fig:exp1-fog-density}, increasing fog density from 0\% to 66.67\% consistently degrades the ensemble's $\mu_{\mathcal{S}}$ across all four adversarial configurations, significantly influencing detection performance. Similarly, the presence of an adversary introducing occlusion consistently delays stop sign classification. Further, analysis of heatmaps, such as Figure \ref{fig:heatmap:experiment1:f-sal}, quantifying sensitivity using $\sigma_{\mathcal{S}}$, highlights that the highest levels, characterized by elevated $\sigma_{\mathcal{S}}$ values, occur under combinations of high fog density and low sun altitude angles (representing low-light conditions). These findings confirm these specific scenarios as particularly challenging for perception robustness.

For the remainder of the experiment set, we evaluated $\mu_{\mathcal{S}}$ and $\sigma_{\mathcal{S}}$ using our ensemble of perception models for every distance in each experiment. While ensemble performance varied throughout the experiment set, we single-out a handful of experiments that demonstrate a strong relationship between certain subsets of  factors and degraded perception performance (based on $\mu_{\mathcal{S}}$ and $\sigma_{\mathcal{S}}$). The following case studies illustrate these findings, demonstrating how specific environmental and adversarial conditions induce substantial variability into the perception system.\\

\subsubsection{Case Study 1: Delayed Detection and High Predictive Sensitivity due to Moderate Occlusion}
\label{sssec:results:case_study1}

\begin{figure*}[!ht]
    \centering
    \subfloat[CARLA simulation view of Experiment 117, illustrating the moderate occlusion of the stop sign by a parked ambulance.]{\includegraphics[width=0.48\linewidth, valign=m, margin=0cm .9cm]{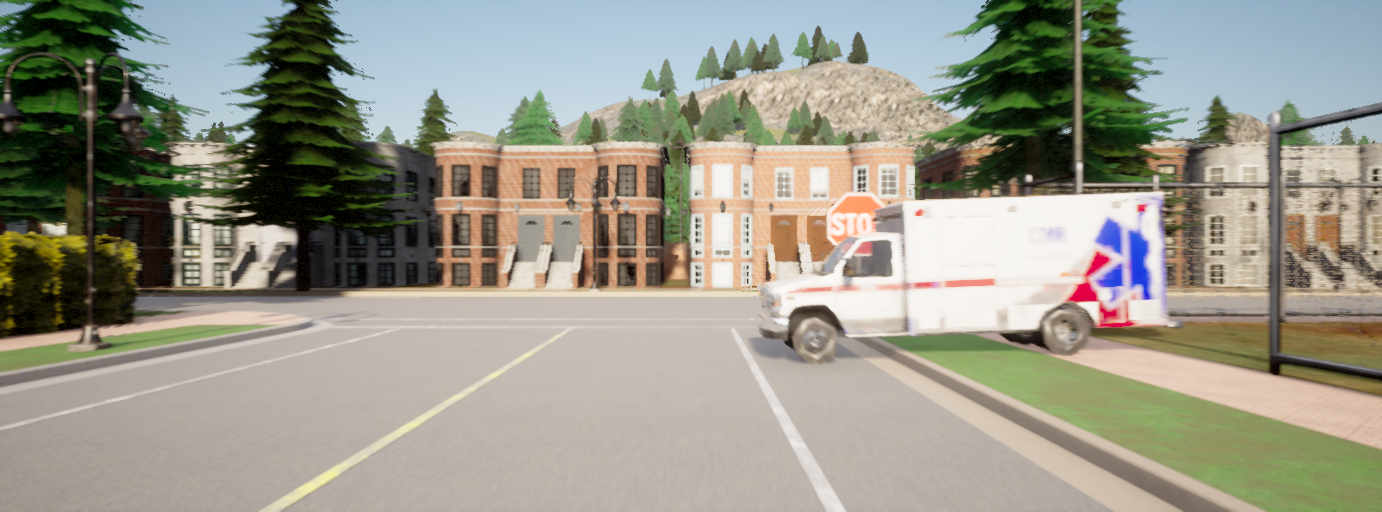}\label{fig:exp1_117c}}%
    \hfill
    \subfloat[Mean prediction probability ($\mu_{\mathcal{S}}$) and standard deviation ($\sigma_{\mathcal{S}}$) for stop sign detection. The vertical red dotted line indicates the stopping distance ($sd = 25.55$ m), and the horizontal green dotted line marks the confidence threshold ($\theta_{\mathcal{S}} = 0.75$). The gray shaded area represents $\sigma_{\mathcal{S}}$.]{\includegraphics[width=0.48\linewidth, valign=m]{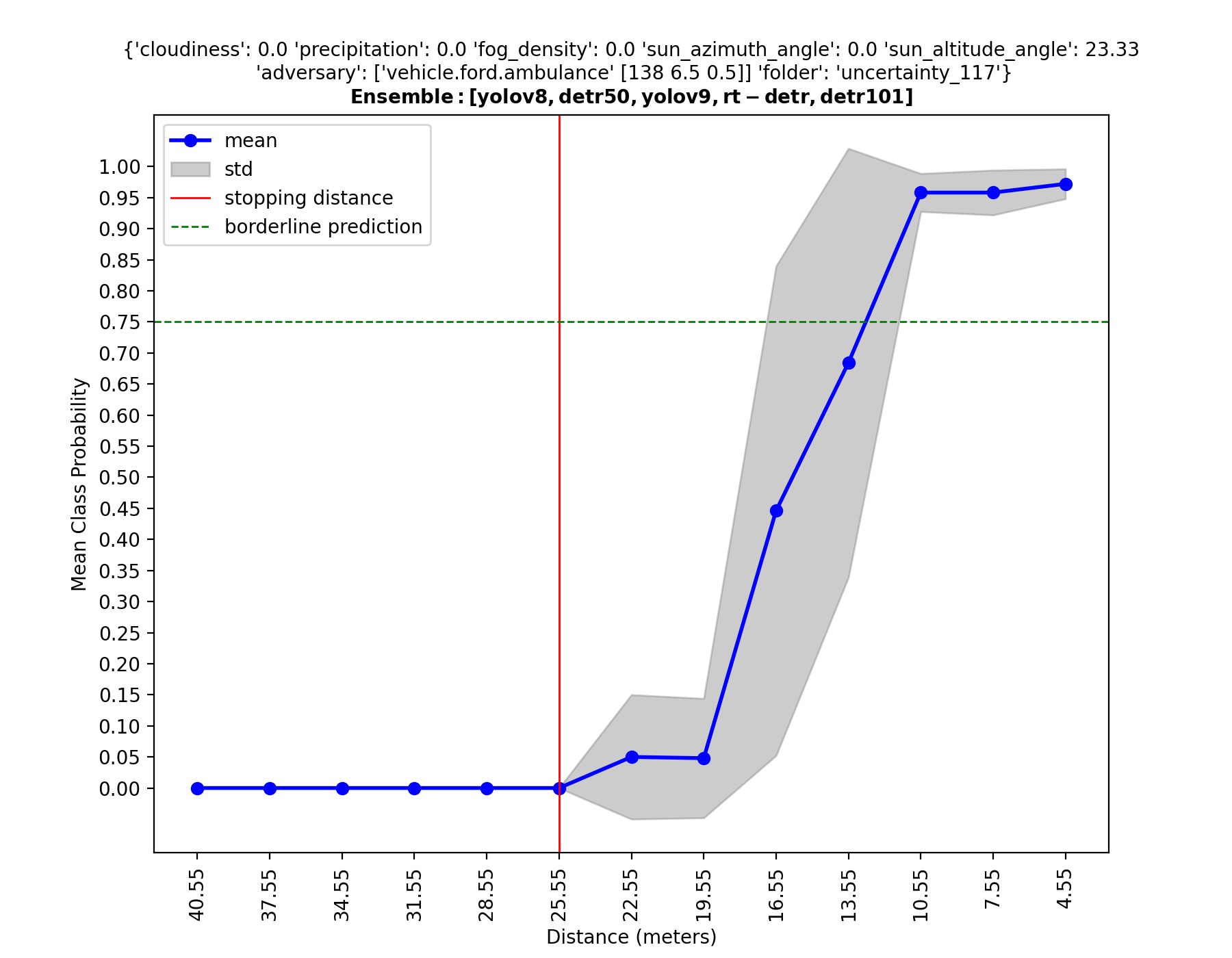}\label{fig:exp1_117g}}
    \caption{Case Study 1: Perception performance and visual context under moderate occlusion. This scenario demonstrates delayed detection and high predictive sensitivity due as the stop sign is partially obscured by an adversarial vehicle, leading to an unsafe condition.}
    \label{fig:exp1_117_composite}
\end{figure*}

Figure \ref{fig:exp1_117_composite} presents a composite view of Experiment 117, a scenario characterized by optimal weather conditions (0\% cloudiness, precipitation, and fog density, with a sun altitude angle of 23.33 degrees) but critically, the presence of a moderate occlusion. As defined in the methodology, this moderate occlusion is represented by a parked Ford ambulance partially obscuring the stop sign ($\mathcal{S}$), as depicted in Figure \ref{fig:exp1_117c}. This setup was designed to investigate the impact of specific adversarial object types on perception performance.

As illustrated in Figure \ref{fig:exp1_117g}, the ensemble's mean prediction probability ($\mu_{\mathcal{S}}$) for $\mathcal{S}$ remains at or near zero for a substantial distance, demonstrating that all participating models struggled significantly with detection and classification. Crucially, at the stopping distance ($sd = 25.55$ m), $\mu_{\mathcal{S}}$ is effectively zero, indicating a complete failure to achieve the required confidence threshold ($\theta_{\mathcal{S}} = 0.75$) to enter region $\mathcal{R}$ before the vehicle reaches the critical braking point. The ensemble only achieves confidence above $\theta_{\mathcal{S}}$ at a much closer range, specifically below $d_{\mathcal{S}} = 13.55$ m.

Concurrent with this delayed detection, the standard deviation ($\sigma_{\mathcal{S}}$) band exhibits pronounced characteristics of high predictive sensitivity. While $\sigma_{\mathcal{S}}$ is narrow at close distances, it's significantly wider in the critical range between \ $d_{\mathcal{S}} \approx 25.55\mbox{ to }10.55$ m, indicating substantial disagreement among models precisely when confident detection is most needed. This poor performance is directly attributable to the partial occlusion of the stop sign by the parked adversarial vehicle. At 30 MPH speed, this scenario unequivocally leads to a potentially unsafe driving condition, as the vehicle may not be able to stop before the stop sign, creating a potential safety hazard due to the perception system's delayed and uncertain inference.\\

\subsubsection{Case Study 2: Confidence Fluctuation and Loss of Temporal Consistency}
\label{sssec:results:case_study2}

\begin{figure*}[!ht]
    \centering
    \subfloat[CARLA simulation view of Experiment 325, illustrating the moderate occlusion by a parked ambulance under nighttime rainy conditions.]{\includegraphics[width=0.48\linewidth, valign=m, margin=0cm .9cm]{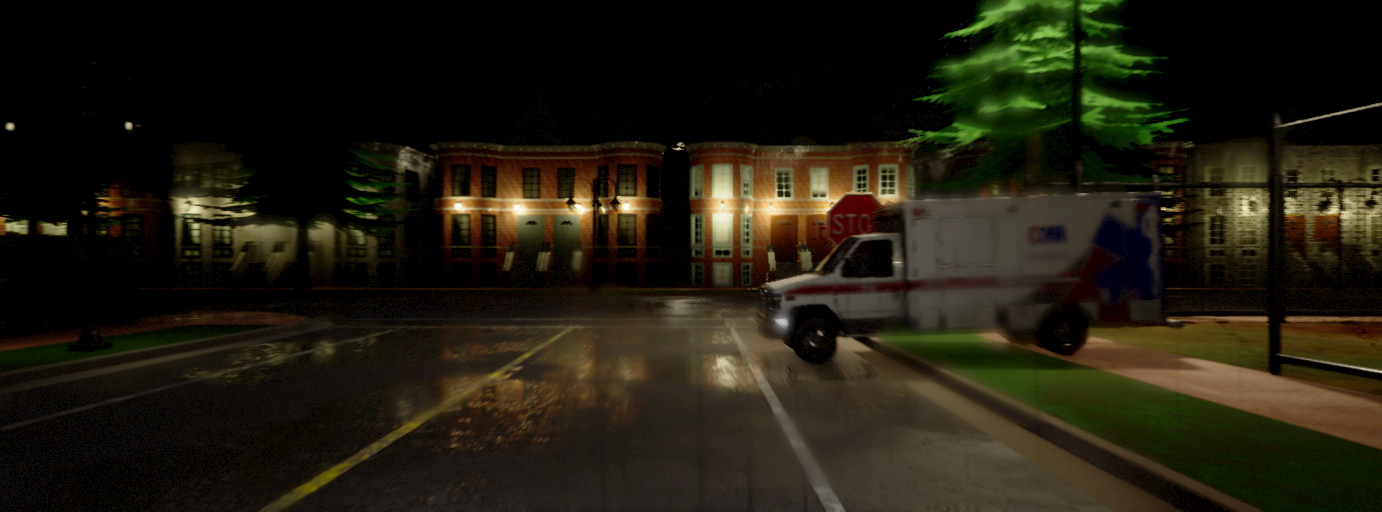}\label{fig:exp1_325c}}
    \hfill
    \subfloat[Mean prediction probability ($\mu_{\mathcal{S}}$) and standard deviation ($\sigma_{\mathcal{S}}$) for stop sign detection. The vertical red dotted line indicates the stopping distance ($sd = 25.55$ m), and the horizontal green dotted line marks the confidence threshold ($\theta_{\mathcal{S}} = 0.75$). The gray shaded area represents $\sigma_{\mathcal{S}}$.]{\includegraphics[width=0.48\linewidth, valign=m]{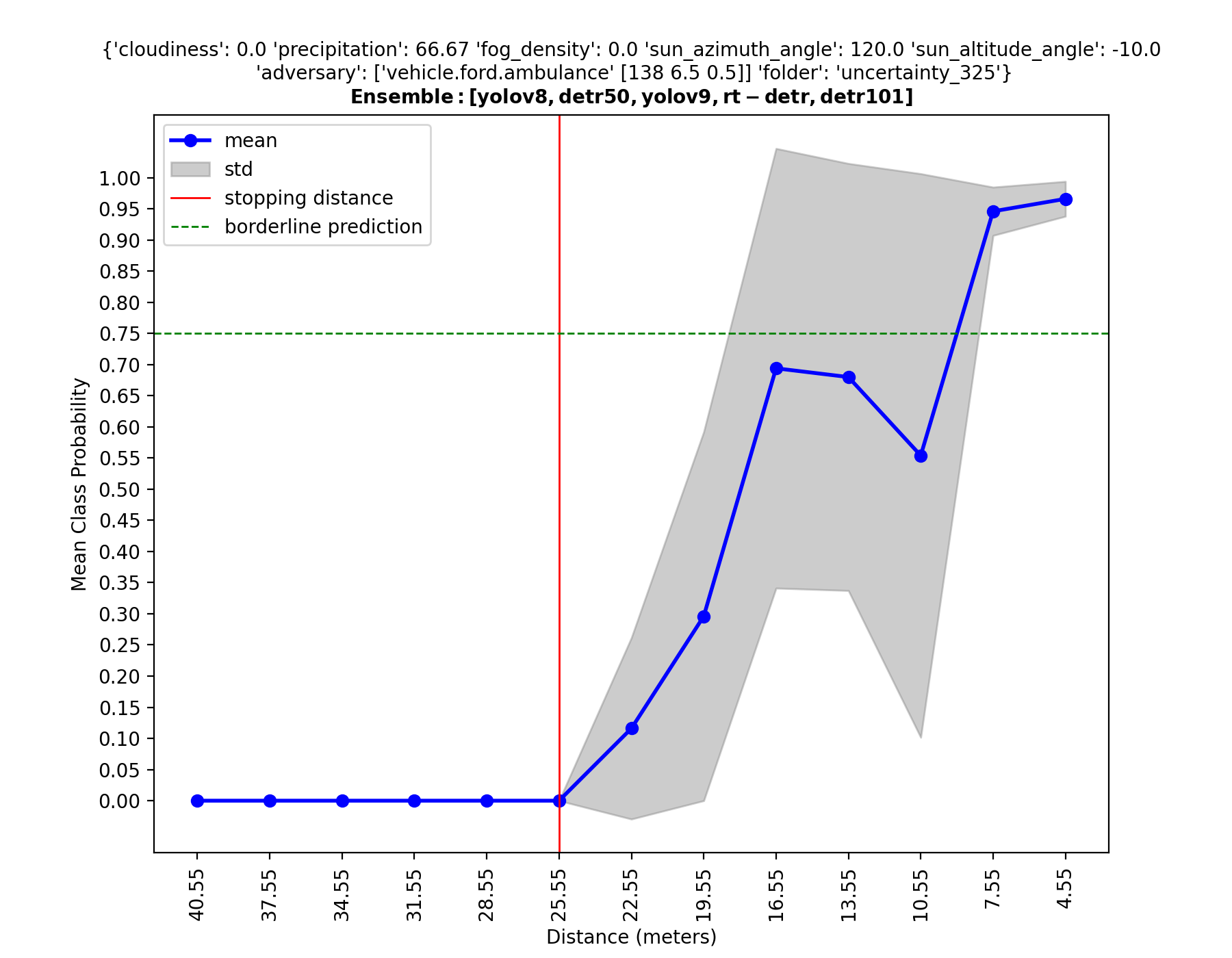}\label{fig:exp1_325g}}
    \caption{Case Study 2: Perception performance and visual context under nighttime rainy conditions. This scenario demonstrates critical confidence fluctuation and loss of temporal consistency, resulting in an unsafe condition.}
    \label{fig:exp1_325_composite}
\end{figure*}

Figure \ref{fig:exp1_325_composite} presents the perception performance in Experiment 325, a scenario that critically demonstrates a violation of the perception system's temporal consistency. This experiment features a high precipitation of 66.67\% and a low sun altitude angle ($-10.0\degree$), indicating a challenging nighttime rainy situation (with no cloudiness or fog). This is combined with the presence of a moderate occlusion shown in Figure \ref{fig:exp1_325c}).

As depicted in Figure \ref{fig:exp1_325g}, the ensemble's mean prediction probability ($\mu_{\mathcal{S}}$) for the stop sign ($\mathcal{S}$) exhibits a distinct and concerning non-monotonic behavior. After an initial slow climb from near zero, $\mu_{\mathcal{S}}$ reaches approximately 0.68 at $d_{\mathcal{S}} = 16.55$ m. However, instead of continuing to increase or maintain confidence as the ego approaches $\mathcal{S}$, $\mu_{\mathcal{S}}$ then drops significantly to approximately 0.55 at $d_{\mathcal{S}} = 13.55$ m, before finally recovering and reaching the confidence threshold ($\theta_{\mathcal{S}} = 0.75$) only around $d_{\mathcal{S}} \approx 12.55$ m. Crucially, at $sd = 25.55$ m, $\mu_{\mathcal{S}}$ remains negligible. Crucially, at the stopping distance ($sd = 25.55$ m), $\mu_{\mathcal{S}}$ remains negligible, indicating a failure to achieve required confidence to enter the safety quadrant ($\mathcal{R}$) by the critical point. This non-monotonic behavior directly violates the desired temporal consistency for robust perception, as confidence drops after an initial gain.

Concurrent with this alarming fluctuation, the standard deviation ($\sigma_{\mathcal{S}}$) band is exceptionally wide in the critical range ($d_{\mathcal{S}} \approx 25.55 \mbox{ to } 10.55$ m). This broad band signifies very high disagreement among the ensemble models, particularly during the period of confidence drop and recovery. This non-monotonic behavior directly violates the temporal consistency requirement as confidence drops after an initial gain, leading to an unreliable perception assessment. This scenario unequivocally results in a potentially unsafe driving condition, severely compromising the ego's ability to react within necessary safety margins due to delayed and unstable perception.\\

\subsubsection{Case Study 3: Severe Perception Performance Degradation from Combined Rain and Occlusion}
\label{sssec:results:case_study3}

\begin{figure*}[!ht]
    \centering
    \subfloat[CARLA simulation view of Experiment 330, illustrating the moderate occlusion by a parked ambulance under high precipitation.]{\includegraphics[width=0.48\linewidth, valign=m, margin=0cm .9cm]{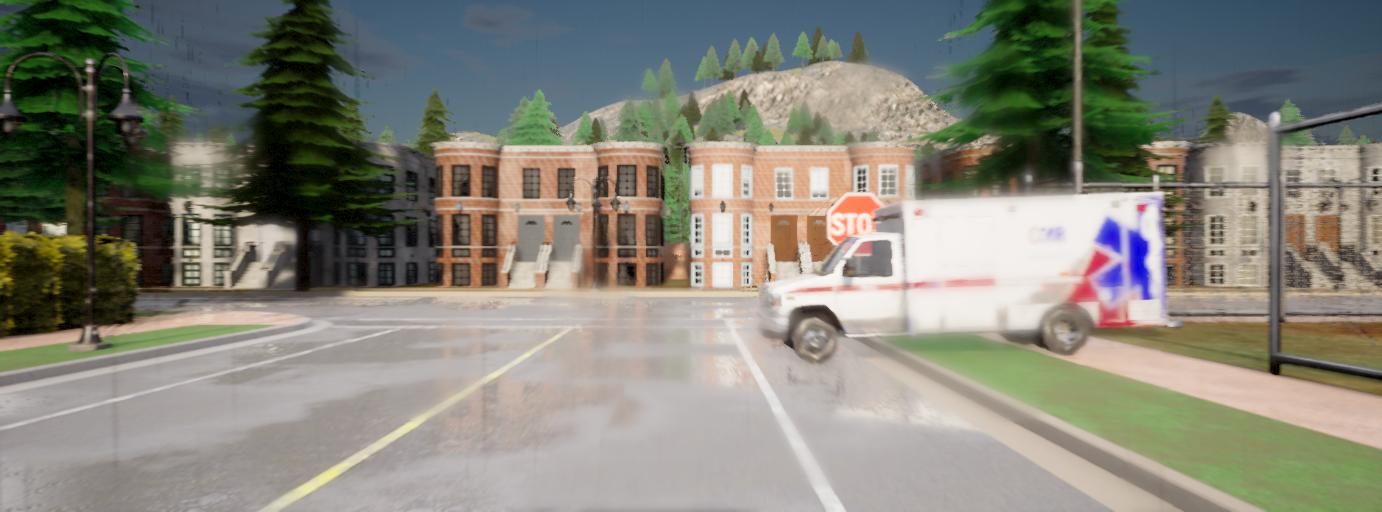}\label{fig:exp1_330c}}
    \hfill
    \subfloat[Mean prediction probability ($\mu_{\mathcal{S}}$) and standard deviation ($\sigma_{\mathcal{S}}$) for stop sign detection. The vertical red dotted line indicates the stopping distance ($sd = 25.55$ m), and the horizontal green dotted line marks the confidence threshold ($\theta_{\mathcal{S}} = 0.75$). The gray shaded area represents $\sigma_{\mathcal{S}}$.]{\includegraphics[width=0.48\linewidth, valign=m]{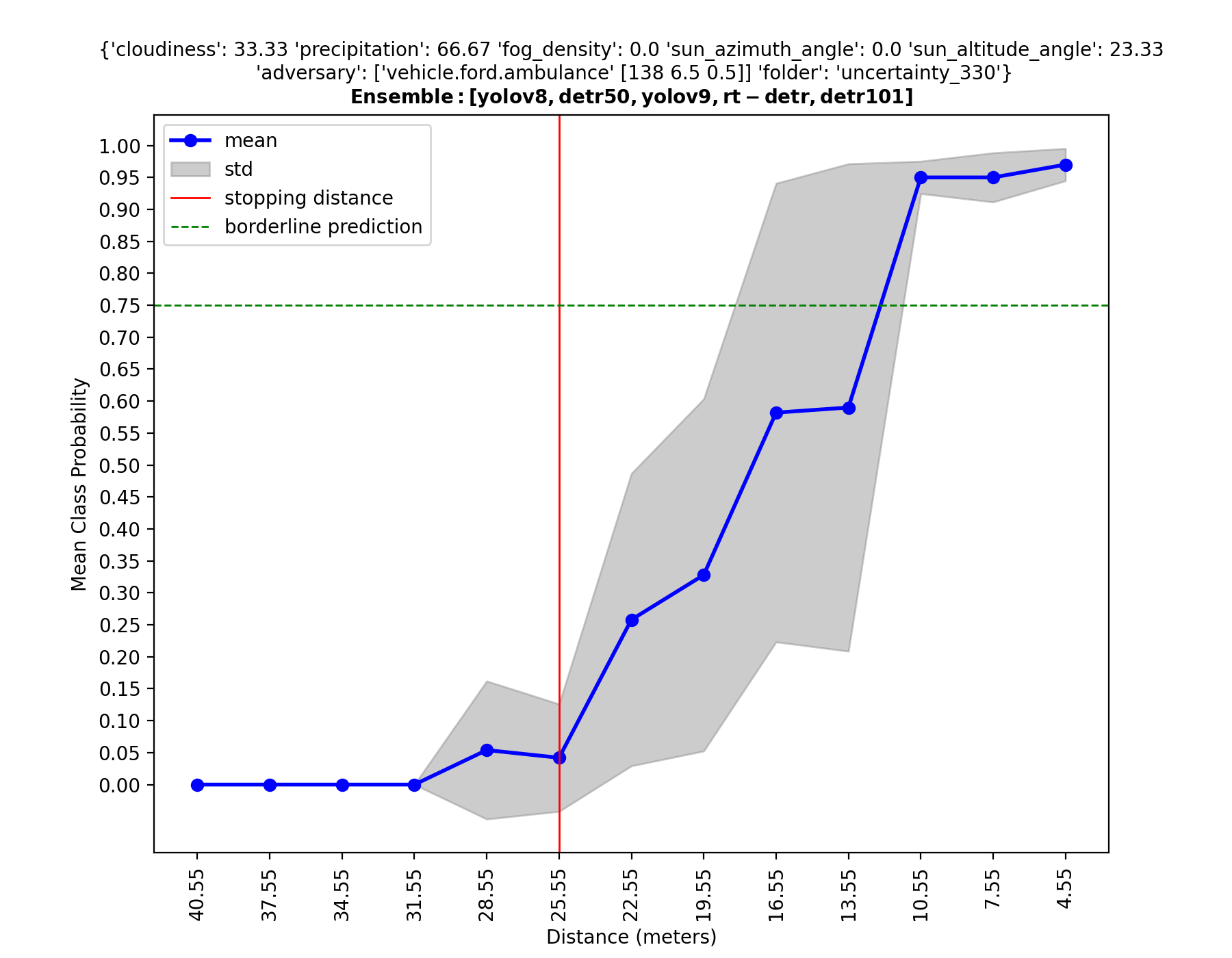}\label{fig:exp1_330g}}
    \caption{Case Study 3: Perception performance and visual context under high precipitation and moderate occlusion. This scenario demonstrates severe predictive sensitivity and delayed detection, leading to an unsafe condition.}
    \label{fig:exp1_330_composite}
\end{figure*}

The perception performance in Experiment 330, presented in Figure \ref{fig:exp1_330_composite}, exemplifies a scenario with challenging environmental conditions and an adversarial object. This experiment features a moderate cloudiness of 33.33\% and a notably high precipitation of 66.67\% (with 0\% fog density), characterizing a distinctly rainy daytime environment (sun altitude angle of 23.33 degrees). This is combined with the presence of a moderate occlusion as shown in Figure \ref{fig:exp1_330c}).

A close examination of Figure \ref{fig:exp1_330g} shows that the ensemble's mean prediction probability ($\mu_{\mathcal{S}}$) for $\mathcal{S}$ remains near zero for a substantial distance. Notably, $\mu_{\mathcal{S}}$ exhibits a step-like increment as \textit{ego} approaches $\mathcal{S}$ after the stopping distance ($sd = 25.55$ m). This behavior suggests that ensemble models gain confidence in discrete, sudden bursts rather than a smooth, continuous increase. At $sd = 25.55$ m, $\mu_{\mathcal{S}}$ is critically low (approximately 0.05), indicating a clear failure to meet the confidence threshold ($\theta_{\mathcal{S}} = 0.75$) at the safety-critical point. The ensemble only achieves confidence above $\theta_{\mathcal{S}}$ at a much closer range, $d_{\mathcal{S}} \approx 11 \mbox{ -- } 12$ m.

\begin{figure*}[!ht]
    \centering
    \subfloat[CARLA simulation view of Experiment 689, illustrating the moderate occlusion by a parked ambulance under compound adverse weather conditions (high cloudiness, moderate precipitation, moderate fog, low light).]{\includegraphics[width=0.48\linewidth, valign=m, margin=0cm .9cm]{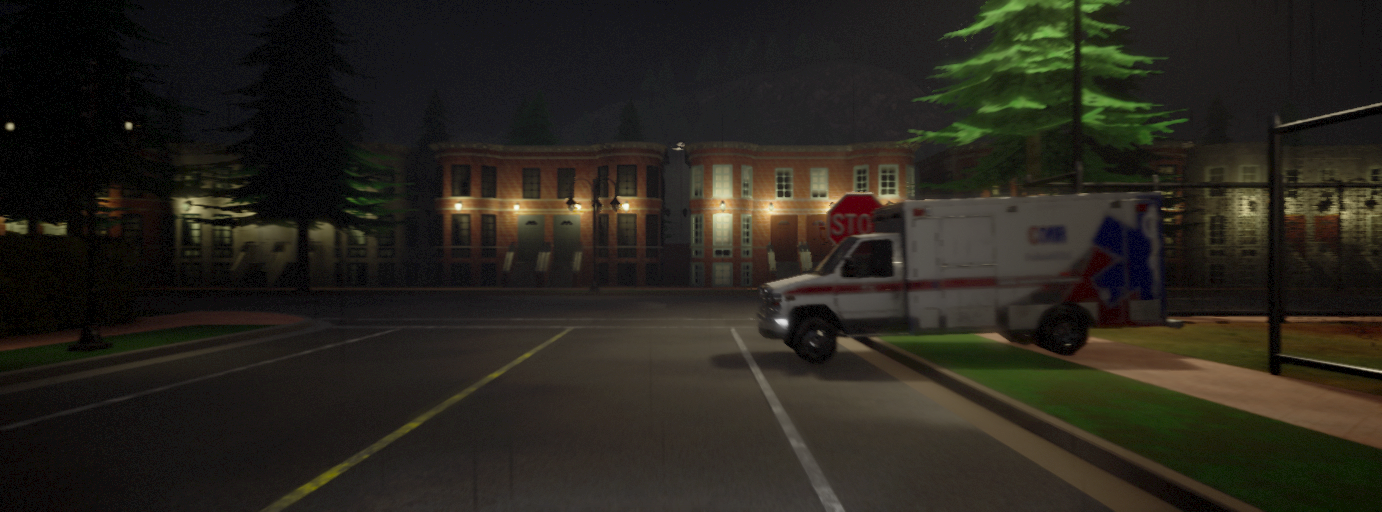}\label{fig:exp1_689c}}
    \hfill
    \subfloat[Mean prediction probability ($\mu_{\mathcal{S}}$) and standard deviation ($\sigma_{\mathcal{S}}$) for stop sign detection. The vertical red dotted line indicates the stopping distance ($sd = 25.55$ m), and the horizontal green dotted line marks the confidence threshold ($\theta_{\mathcal{S}} = 0.75$). The gray shaded area represents $\sigma_{\mathcal{S}}$.]{\includegraphics[width=0.48\linewidth, valign=m]{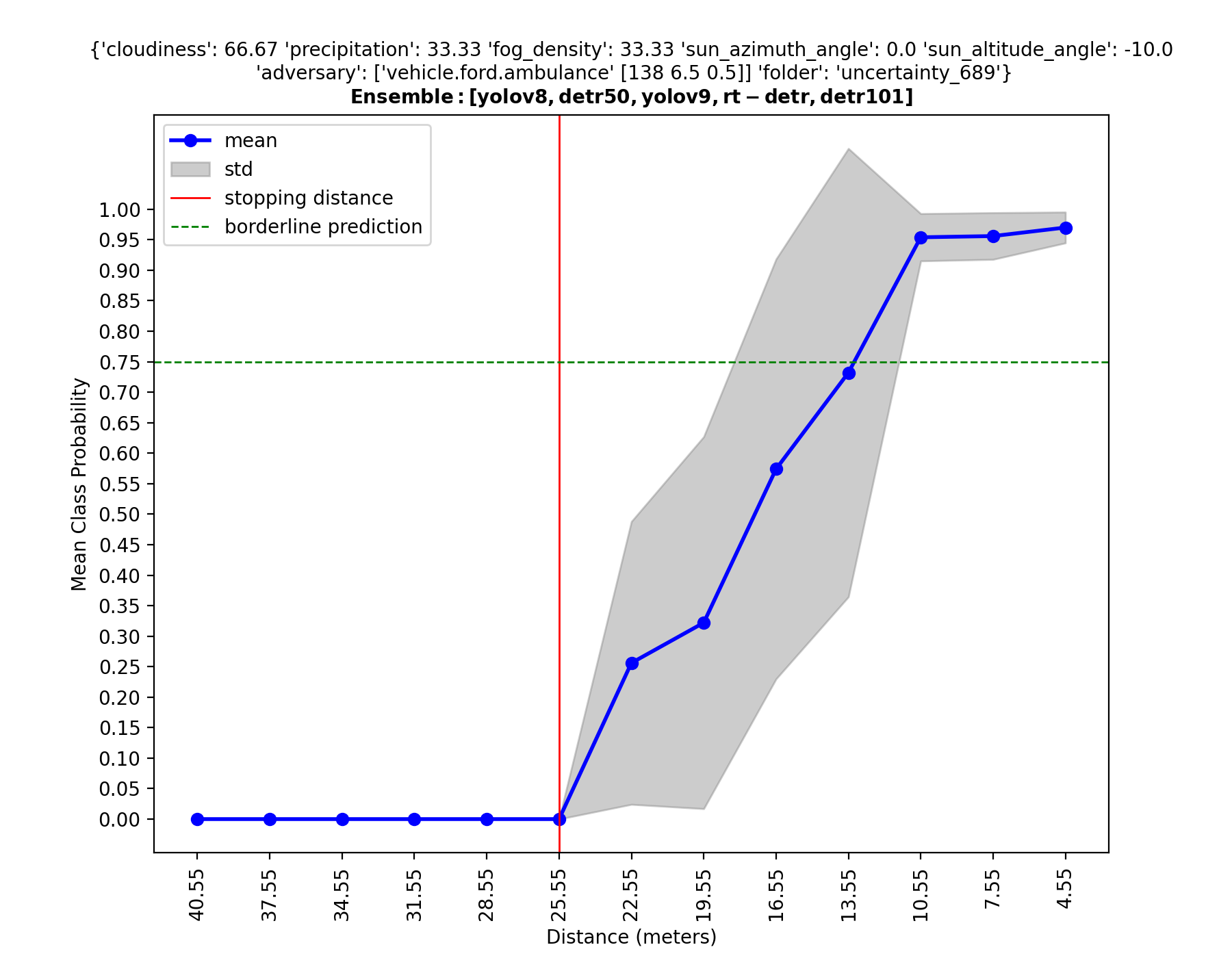}\label{fig:exp1_689g}}

    \caption{Case Study 4: Perception performance and visual context under compound environmental challenges and moderate occlusion. This scenario demonstrates profound perception degradation leading to severely delayed detection and resulting in an unsafe condition.}
    \label{fig:exp1_689_composite}
\end{figure*}

This significant delay in detection and the step-like confidence gain are accompanied by an exceptionally wide spread in the standard deviation ($\sigma_{\mathcal{S}}$) band within the critical range ($d_{\mathcal{S}} \approx 25.55 \mbox{ to } 10.55$ m). This broad range signifies profound disagreement among the ensemble models. Individual models likely alternate between very low confidence and sudden, potentially spurious, detections. This poor performance is directly attributable to the combined effect of the challenging rainy conditions (high precipitation and cloudiness) and the partial occlusion by the adversarial vehicle. Such a scenario unequivocally leads to an unsafe driving condition, as the perception system's confidence is severely compromised and critically delayed.\\

\subsubsection{Case Study 4: Compound Environmental Challenges Leading to Extreme Degradation of Perception Performance}
\label{sssec:results:case_study4}

Experiment 689, visualized in Figure \ref{fig:exp1_689_composite}, represents a scenario with a compounding of multiple environmental challenges and an adversarial object. This experiment features high cloudiness (66.67\%), moderate precipitation (33.33\%), and moderate fog density (33.33\%), combined with a low sun altitude angle ($-10.0$ degrees), indicative of significant low-light/nighttime conditions. A moderate occlusion as shown in Figure \ref{fig:exp1_689c}) is also present.

Analysis of Figure \ref{fig:exp1_689g} reveals that the ensemble's mean prediction probability ($\mu_{\mathcal{S}}$) for the stop sign ($\mathcal{S}$) remains at or near zero for extended distances. At the stopping distance ($sd = 25.55$ m), $\mu_{\mathcal{S}}$ is negligible, indicating a complete failure to meet the confidence threshold ($\theta_{\mathcal{S}} = 0.75$) by the critical safety point. The ensemble only achieves confidence above $\theta_{\mathcal{S}}$ at a much closer range, $d_{\mathcal{S}} \approx 11 \mbox{ -- } 12$ m. This represents a significant delay in reliable detection.

This severe delay is accompanied by an exceptionally wide and persistent spread in the standard deviation ($\sigma_{\mathcal{S}}$) band across the critical range ($d_{\mathcal{S}} \approx 25.55 \mbox{ to } 10.55$ m). This indicates profound disagreement among ensemble models  throughout the crucial detection phase. The combined adverse effects of high cloudiness, moderate precipitation, moderate fog, and low sun altitude, exacerbated by the moderate occlusion, severely degrade the perception system's ability to consistently or confidently detect the stop sign. Consequently, this scenario unequivocally leads to an unsafe driving condition, as the perception system's delayed and highly uncertain inference critically compromises the ego's ability to react within necessary safety margins.

The case studies presented (Experiments 117, 325, 330, and 689) are representative examples from Experiment Set 1 that collectively provide compelling evidence of significant perception degradation under various challenging environmental and adversarial conditions. The sensitivity analysis, performed across numerous scenarios, clearly demonstrates that factors such as moderate occlusion, high fog density, high precipitation, and low sun altitude angle (indicative of low-light conditions) consistently lead to delayed detection, fluctuations in mean prediction probability ($\mu_{\mathcal{S}}$), and markedly wide standard deviation ($\sigma_{\mathcal{S}}$) bands. These findings underscore the direct correlation between specific adverse conditions and disagreement among the model in the ensemble's predictions. Furthermore, the compounding effect of multiple challenging factors, as seen in Case Study 4, exacerbates this disagreement, pushing the system into unsafe operating conditions. This phase of experimentation successfully identified the critical state variables that profoundly impact perception robustness, validating the existence of quantifiable predictive sensitivity in real-world relevant scenarios.

\begin{figure*}[!ht]
    \centering
    \subfloat[CARLA simulation view of Experiment 156, illustrating the adversarial object (a tree) partially covering the stop sign under extreme compound weather conditions.]{\includegraphics[width=0.48\linewidth, valign=m, margin=0cm .9cm]{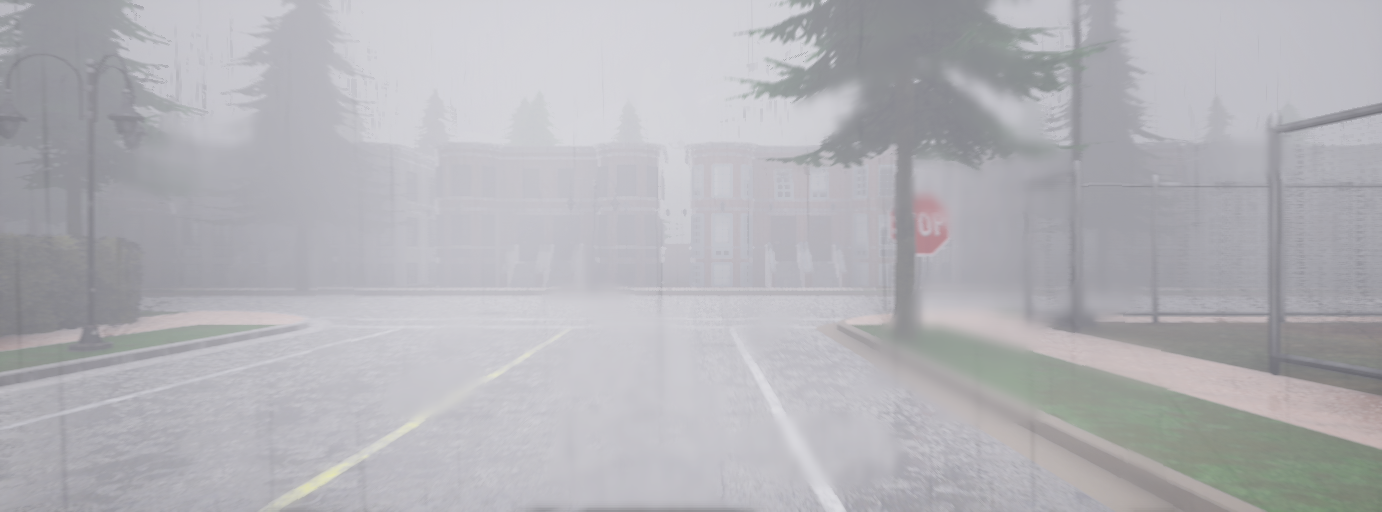}\label{fig:exp2_156c}}%
    \hfill
    \subfloat[Mean prediction probability ($\mu_{\mathcal{S}}$) and standard deviation ($\sigma_{\mathcal{S}}$) for stop sign detection. The vertical red dotted line indicates the stopping distance ($sd = 25.55$ m), and the horizontal green dotted line marks the confidence threshold ($\theta_{\mathcal{S}} = 0.75$). The gray shaded area represents $\sigma_{\mathcal{S}}$.]{\includegraphics[width=0.48\linewidth, valign=m]{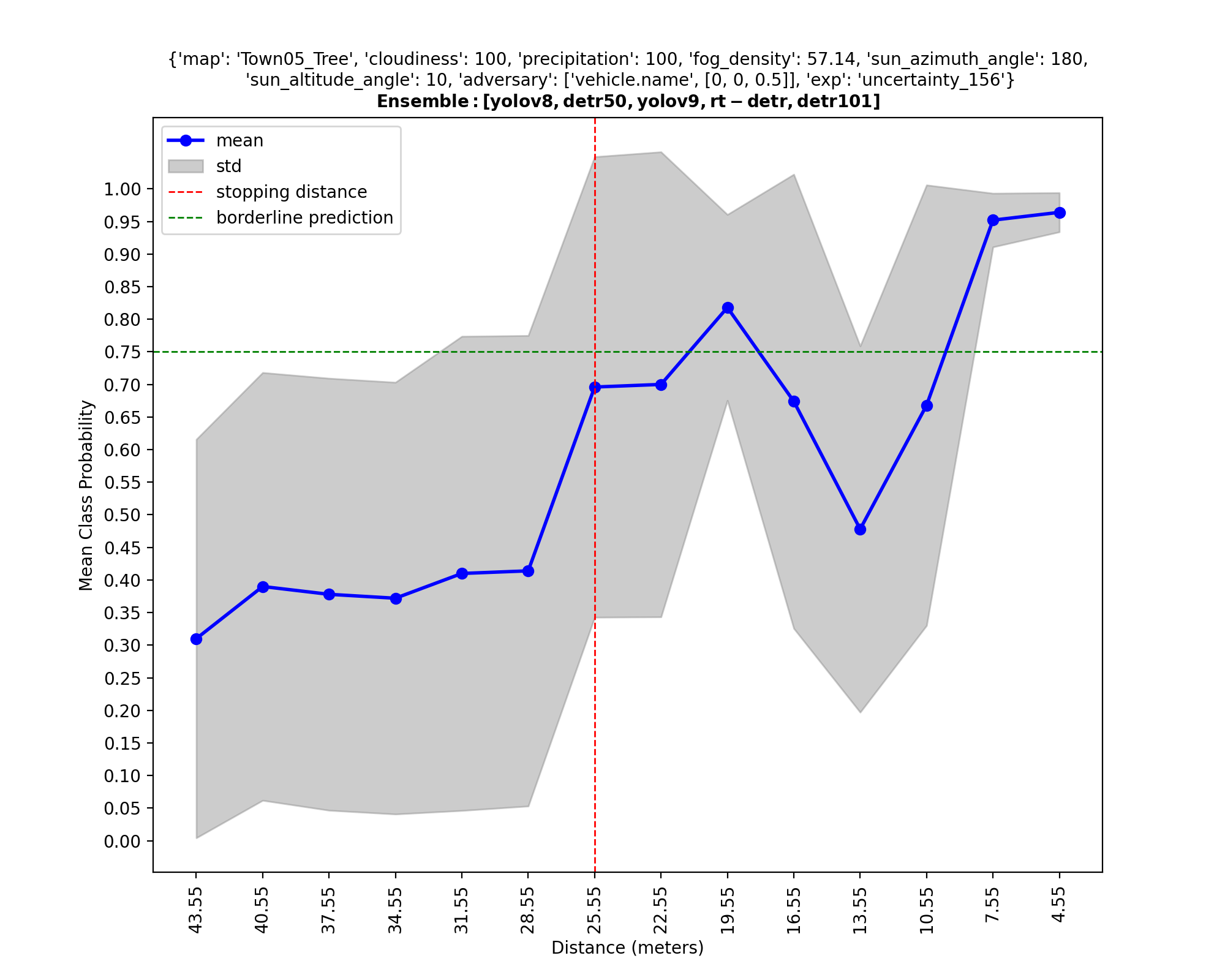}\label{fig:exp2_156g}}
    \caption{Case Study 1 (Experiment Set 2): Perception performance and visual context under compound extreme conditions and occlusion by a tree. This scenario demonstrates catastrophic degradation, extreme predictive sensitivity, and severe loss of temporal consistency, leading to an unsafe condition.}
    \label{fig:exp2_156_composite}
\end{figure*}

\subsection{Robustness Limits and Extreme Sensitivity (Experiment Set 2)}
\label{subsec:results:exp2}


The intensified operating conditions applied in this second experimental set rigorously tested the perception system's limits, consistently revealing instances of extreme model disagreement  and measurable degradation in robustness. Such observations are vital for delineating the perception system's ODD, defining the specific operating conditions under which it can function safely and reliably. The following four case studies, drawn from Experiment Set 2, exemplify how specific combinations of extreme environmental parameters and complex adversarial objects lead to these profound challenges in object detection and robust perception.\\

\subsubsection{Case Study 1: Catastrophic Degradation from Compound Extreme Conditions}
\label{sssec:results:exp2:case_study1}

Figure \ref{fig:exp2_156_composite} illustrates the perception performance in Experiment 156, a scenario demonstrating a severe breakdown in robustness due to a compounding of extreme environmental challenges. This experiment features a full cloud cover (100\% cloudiness), extreme precipitation (100\%), significant fog (57.14\% fog density), and a low sun altitude angle (10 degrees), indicative of dusk-like conditions. These factors are combined with the presence of an adversarial object: a tree partially covering the stop sign ($\mathcal{S}$) and \textit{ego}'s field of view as it drives towards $\mathcal{S}$ (as shown in Figure \ref{fig:exp2_156c}).

As depicted in Figure \ref{fig:exp2_156g}, the ensemble's mean prediction probability ($\mu_{\mathcal{S}}$) for $\mathcal{S}$ exhibits highly erratic and non-monotonic behavior. $\mu_{\mathcal{S}}$ starts low and, crucially, at the stopping distance ($sd = 25.55$ m), it is approximately 0.70, failing to achieve the required confidence ($\theta_{\mathcal{S}} = 0.75$) to enter the safety quadrant ($\mathcal{R}$) at this critical point. The perception system remains outside $\mathcal{R}$ as the ego approaches $\mathcal{S}$. Even after $sd$ (i.e., when the vehicle is already within the stopping distance), $\mu_{\mathcal{S}}$ shows highly unstable behavior: it briefly climbs above $\theta_{\mathcal{S}}$ at $d_{\mathcal{S}} = 19.55$ m, but then immediately drops significantly below $\theta_{\mathcal{S}}$, fluctuating wildly before finally recovering to sustained high confidence at very close ranges ($d_{\mathcal{S}} \approx 8$ m). This represents a severe failure of temporal consistency within the critical operating range.

Concurrent with this erratic $\mu_{\mathcal{S}}$ trend, the standard deviation ($\sigma_{\mathcal{S}}$) band is exceptionally wide and pervasive throughout the entire range of distances until very close proximity. This consistently broad band signifies profound and persistent disagreement among the ensemble models, around most of the crucial detection phase. This scenario unequivocally results in an unsafe driving condition, as the perception system utterly fails to provide reliable or consistent input within the necessary safety margins, indicating operation far outside its defined ODD.\\

\subsubsection{Case Study 2: Extreme Predictive Sensitivity from Adversarial Stop Sign Alteration}
\label{sssec:results:exp2:case_study2}

\begin{figure*}[!ht]
    \centering
    \subfloat[CARLA simulation view of Experiment 438, illustrating the adversarially altered stop sign under optimal weather conditions.]{\includegraphics[width=0.48\linewidth, valign=m, margin=0cm .9cm]{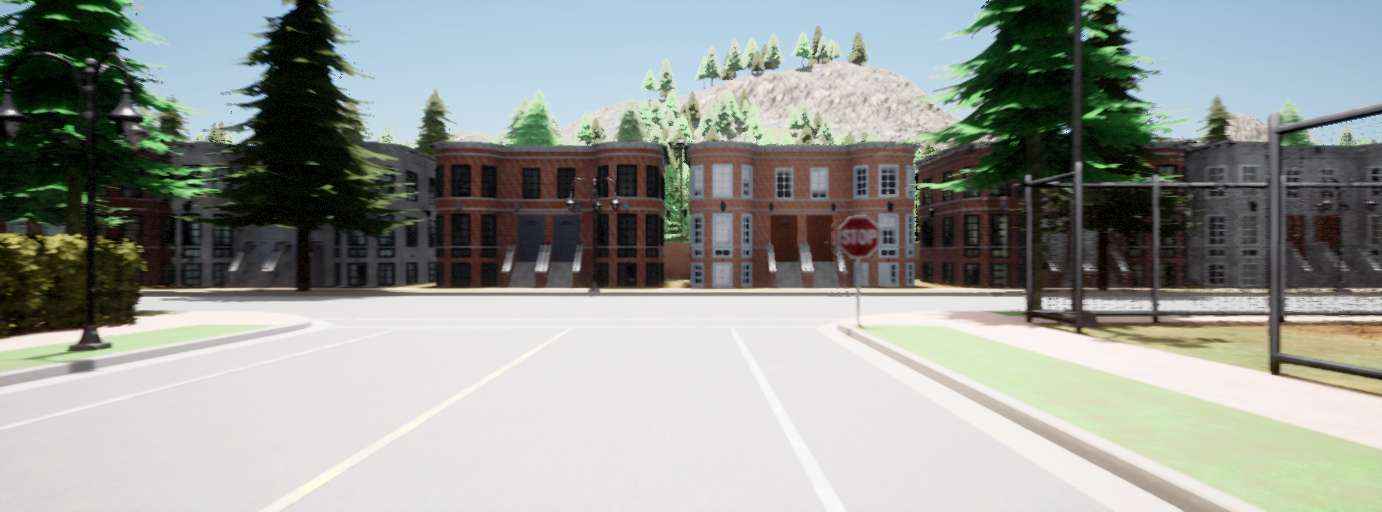}\label{fig:exp2_438c}}%
    \hfill
    \subfloat[Mean prediction probability ($\mu_{\mathcal{S}}$) and standard deviation ($\sigma_{\mathcal{S}}$) for stop sign detection. The vertical red dotted line indicates the stopping distance ($sd = 25.55$ m), and the horizontal green dotted line marks the confidence threshold ($\theta_{\mathcal{S}} = 0.75$). The gray shaded area represents $\sigma_{\mathcal{S}}$.]{\includegraphics[width=0.48\linewidth, valign=m]{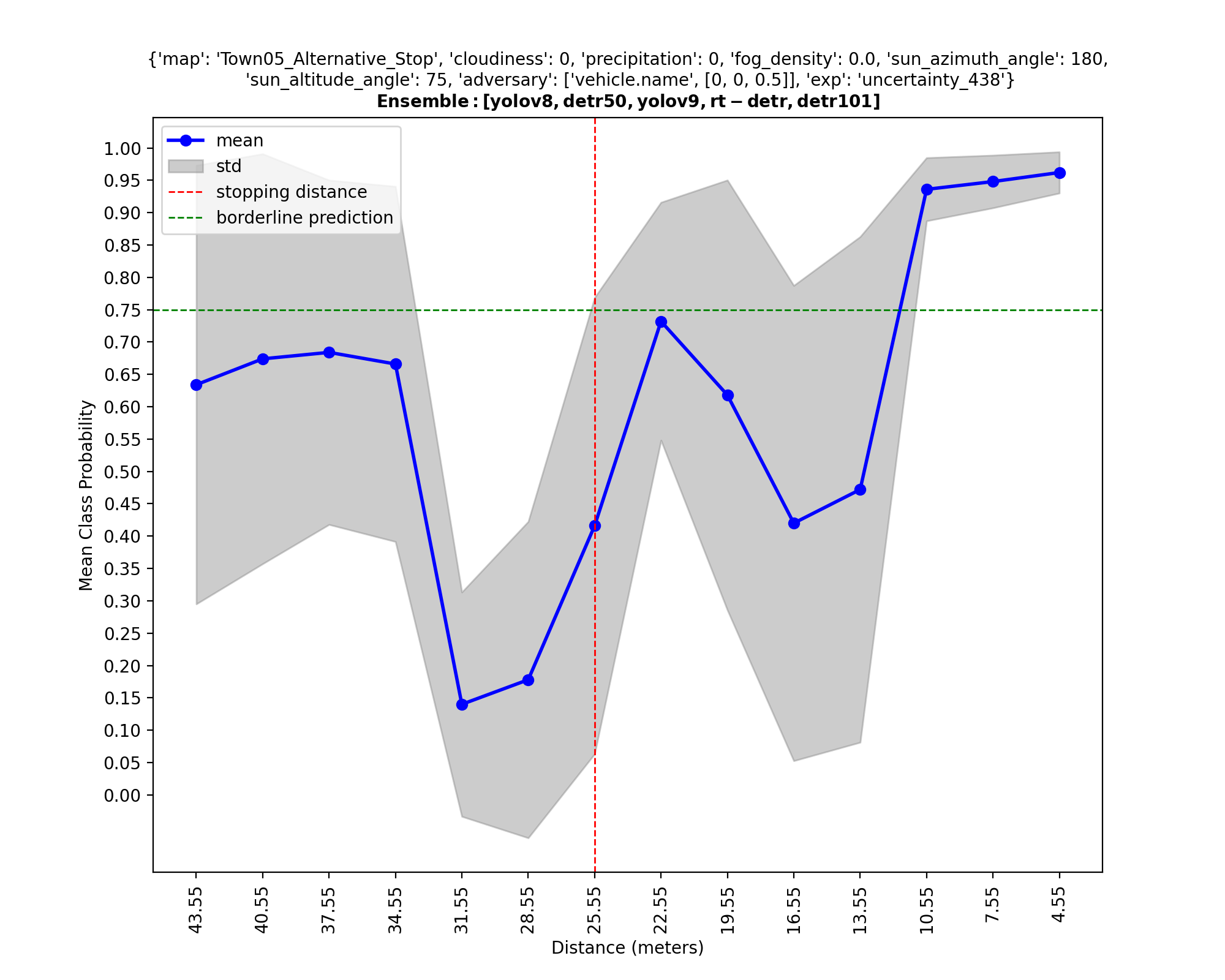}\label{fig:exp2_438g}}
    \caption{Case Study 2 (Experiment Set 2): Perception performance and visual context under adversarial stop sign alteration. This scenario demonstrates extreme  predictive sensitivity, catastrophic degradation, and severe loss of temporal consistency, resulting in an unsafe condition.}
    \label{fig:exp2_438_composite}
\end{figure*}

Figure \ref{fig:exp2_438_composite} illustrates the perception performance in Experiment 438, a scenario that highlights a profound vulnerability of the perception system to direct adversarial manipulation. This experiment was conducted under optimal, clear weather conditions (0\% cloudiness, 0\% precipitation, 0\% fog density, and a high sun altitude angle of 75 degrees) leading to a well-illuminated environment. Critically, however, the stop sign ($\mathcal{S}$) itself was adversarially altered by changing its contrast, texture, and color (as shown in Figure \ref{fig:exp2_438c}).

\begin{figure*}[!ht]
    \centering
    \subfloat[CARLA simulation view of Experiment 690, illustrating the adversarially altered stop sign under severe sun glare conditions.]{\includegraphics[width=0.48\linewidth, valign=m, margin=0cm .9cm]{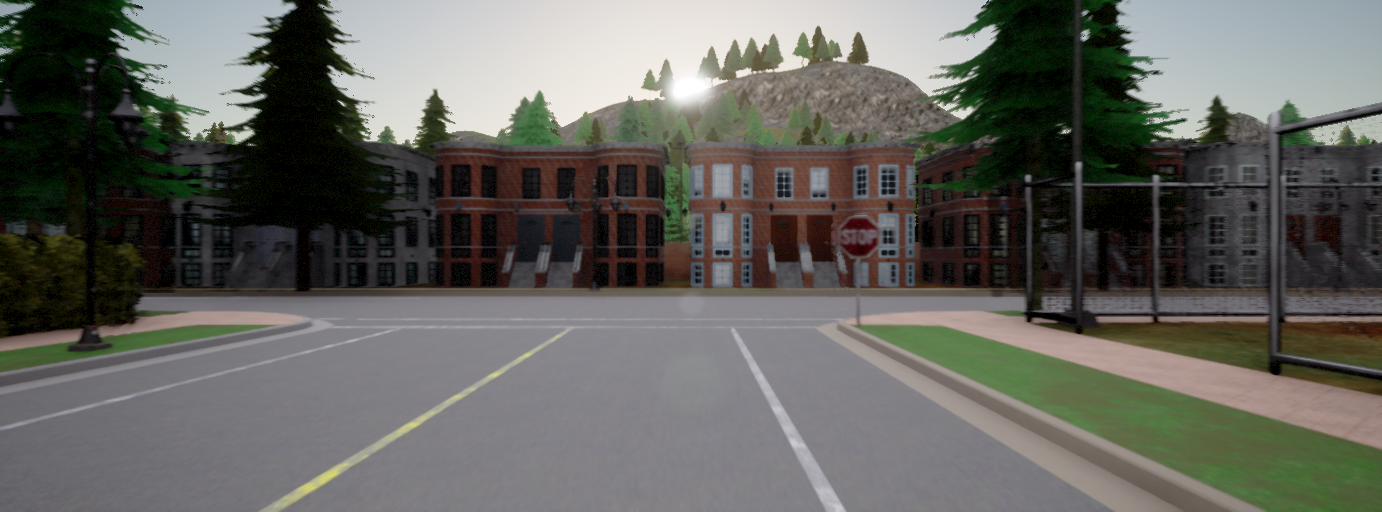}\label{fig:exp2_690c}}
    \hfill
    \subfloat[Mean prediction probability ($\mu_{\mathcal{S}}$) and standard deviation ($\sigma_{\mathcal{S}}$) for stop sign detection. The vertical red dotted line indicates the stopping distance ($sd = 25.55$ m), and the horizontal green dotted line marks the confidence threshold ($\theta_{\mathcal{S}} = 0.75$). The gray shaded area represents $\sigma_{\mathcal{S}}$.]{\includegraphics[width=0.48\linewidth, valign=m]{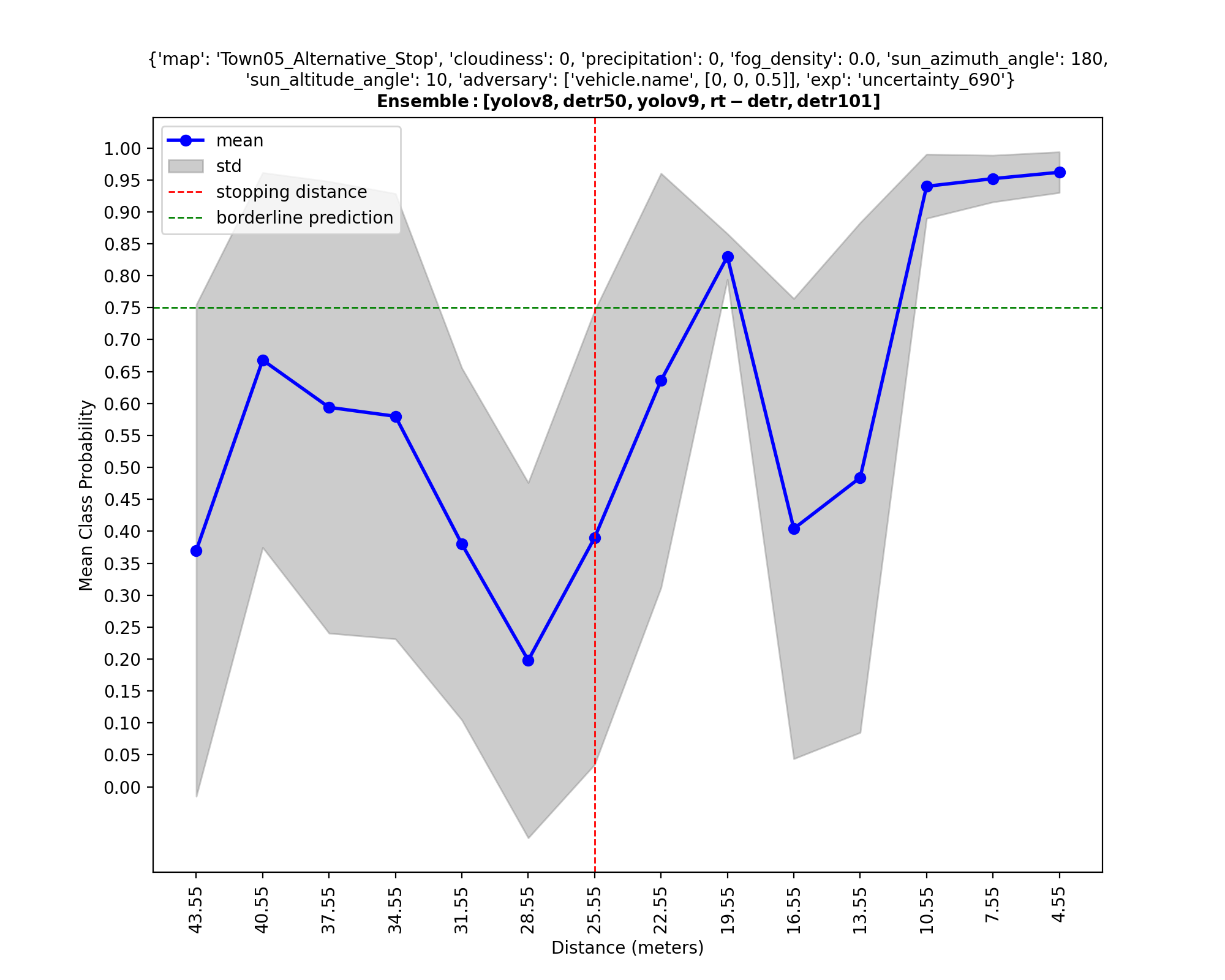}\label{fig:exp2_690g}}
    \caption{Case Study 3 (Experiment Set 2): Perception performance and visual context under adversarial stop sign alteration and severe sun glare. This scenario demonstrates catastrophic predictive sensitivity, extreme degradation, and severe loss of temporal consistency, leading to an unsafe condition.}
    \label{fig:exp2_690_composite}
\end{figure*}

As depicted in Figure \ref{fig:exp2_438g}, the ensemble's mean prediction probability ($\mu_{\mathcal{S}}$) for $\mathcal{S}$ exhibits exceptionally volatile and non-monotonic behavior across the entire detection range. Initially, $\mu_{\mathcal{S}}$ is around 0.65 from $d_{\mathcal{S}} = 43.55 \rightarrow 34.55$ m, accompanied by a wide standard deviation ($\sigma_{\mathcal{S}}$) band, indicating high initial model disagreement. After the $d_{\mathcal{S}} = 34.55$ m mark, $\mu_{\mathcal{S}}$ sees a steep drop to $\approx 0.15 \mbox{ at } d_{\mathcal{S}} = 31.55$ m. Notably, this drop is accompanied by a narrowing of the $\sigma_{\mathcal{S}}$ band, indicating that the models collectively agreed on a very low confidence in detection. From $d_{\mathcal{S}} = 28.55 \rightarrow 22.55$ m, $\mu_{\mathcal{S}}$ shows a sharp increase, but, crucially, at the stopping distance ($sd = 25.55$ m), it is approximately 0.40, representing a clear failure to achieve the required confidence ($\theta_{\mathcal{S}} = 0.75$) to enter region $\mathcal{R}$. The perception system remains outside $\mathcal{R}$ as the ego approaches $\mathcal{S}$. Even after $sd$, $\mu_{\mathcal{S}}$ shows highly unstable behavior with multiple severe drops (e.g., from $\approx 0.70 \mbox{ at } d_{\mathcal{S}} = 22.55$ m to $\approx 0.60 \mbox{ at } 19.55$ m, and further to $\approx 0.42 \mbox{ at } 16.55$ m), before finally recovering to sustained high confidence at very close ranges ($d_{\mathcal{S}} \approx 8$ m). This pattern represents a severe failure of temporal consistency within the critical operating range.

This extreme erraticism, coupled with the persistent inability to provide stable high confidence, unequivocally results in an unsafe driving condition, as the perception system utterly fails to provide reliable or consistent input within the necessary safety margins, demonstrating a critical vulnerability to adversarial alterations. \\

\subsubsection{Case Study 3: Catastrophic Perception Performance Degradation from Adversarial Alteration and Sun Glare}
\label{sssec:results:exp2:case_study3}

Experiment 690 (Figure \ref{fig:exp2_690_composite}) presents a challenging scenario designed to test the system's robustness against compounded visual degradations. Here, optimal weather conditions (0\% cloudiness, 0\% precipitation, 0\% fog density) are combined with an adversarially altered stop sign ($\mathcal{S}$), featuring changed contrast, texture, and color. Adding to this challenge, a low sun altitude angle ($-10.0$ degrees) and a sun azimuth of $180^\circ$ position the sun directly in front of the ego, as vividly shown in Figure \ref{fig:exp2_690c}. This creates intense glare and spectral deflection, significantly obscuring the scene and contributing to dusk-like conditions.

The ensemble's mean prediction probability ($\mu_{\mathcal{S}}$) for $\mathcal{S}$ exhibits exceptionally volatile and non-monotonic behavior across the entire detection range (Figure \ref{fig:exp2_690g}). From far distances, $\mu_{\mathcal{S}}$ starts low (e.g., $\approx 0.65 \mbox{ at } d_{\mathcal{S}} = 40.55$ m) then drops (to $\approx 0.59 \mbox{ at } 37.55$ m), followed by a steep decline to $\approx 0.20 \mbox{ at } 28.55$ m. At the stopping distance ($sd = 25.55$ m), $\mu_{\mathcal{S}}$ is $\approx 0.40$, signifying a clear failure to achieve the required confidence ($\theta_{\mathcal{S}} = 0.75$) to enter region $\mathcal{R}$. The perception system consequently remains outside $\mathcal{R}$ as the ego approaches $\mathcal{S}$. Beyond $sd$, $\mu_{\mathcal{S}}$ continues its highly unstable trend: it briefly climbs (to $\approx 0.82 \mbox{ at } 19.55$~m) then suffers a severe drop (to $\approx 0.40 \mbox{ at } 16.55$~m), before finally recovering to sustained high confidence only at very close ranges ($d_{\mathcal{S}} \approx 8$~m). This pattern unequivocally represents a catastrophic failure of temporal consistency within the critical operating range.

This extreme erraticism is accompanied by an standard deviation ($\sigma_{\mathcal{S}}$) band that is exceptionally wide and pervasive throughout almost the entire range of distances until very close proximity. This consistently broad band signifies profound and persistent disagreement among ensemble models around most of the crucial detection phase. The experiment highlights that even small perturbations in sun altitude angle can profoundly affect the perception system's state. As previously established in Experiment Set 1, sun altitude angle plays a significant role in determining ensemble performance, which is further verified here. This scenario unequivocally results in an unsafe driving condition, as the perception system utterly fails to provide reliable or consistent input within the necessary safety margins. Its operation clearly falls far outside the defined ODD, underscoring extreme vulnerability to combined adversarial alteration and severe environmental glare.

\begin{figure*}[!h]
    \centering
    \subfloat[CARLA simulation view of Experiment 842, illustrating the adversarially altered stop sign under severe fog and low light conditions.]{\includegraphics[width=0.48\linewidth, valign=m, margin=0cm .9cm]{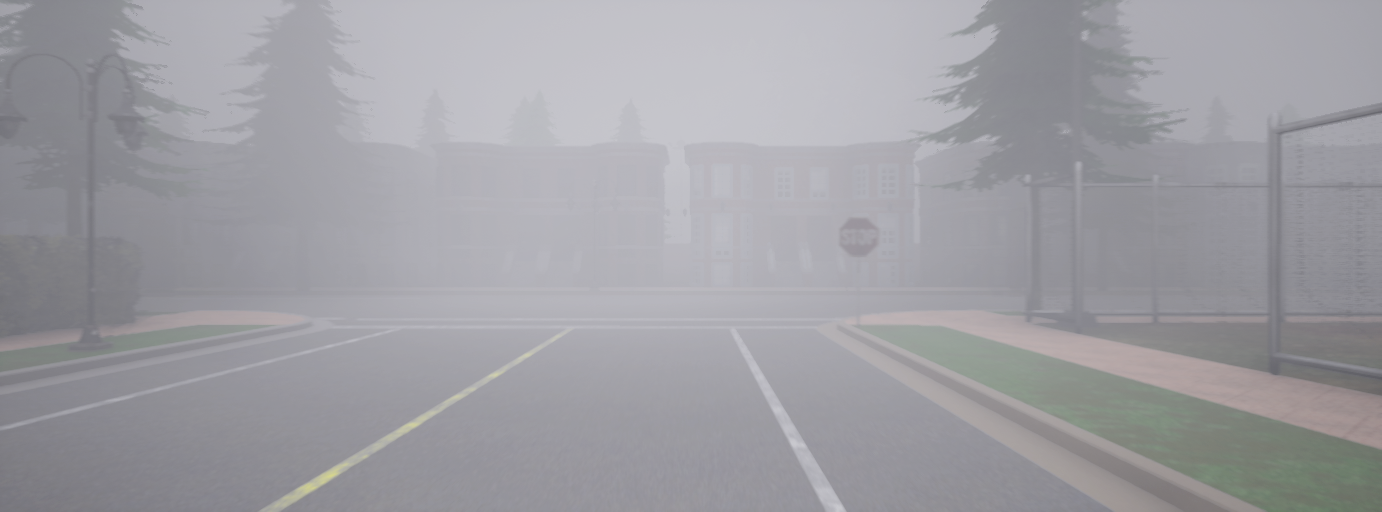}\label{fig:exp2_842c}}
    \hfill 
    \subfloat[Mean prediction probability ($\mu_{\mathcal{S}}$) and standard deviation ($\sigma_{\mathcal{S}}$) for stop sign detection. The vertical red dotted line indicates the stopping distance ($sd = 25.55$ m), and the horizontal green dotted line marks the confidence threshold ($\theta_{\mathcal{S}} = 0.75$). The gray shaded area represents $\sigma_{\mathcal{S}}$.]{\includegraphics[width=0.48\linewidth, valign=m]{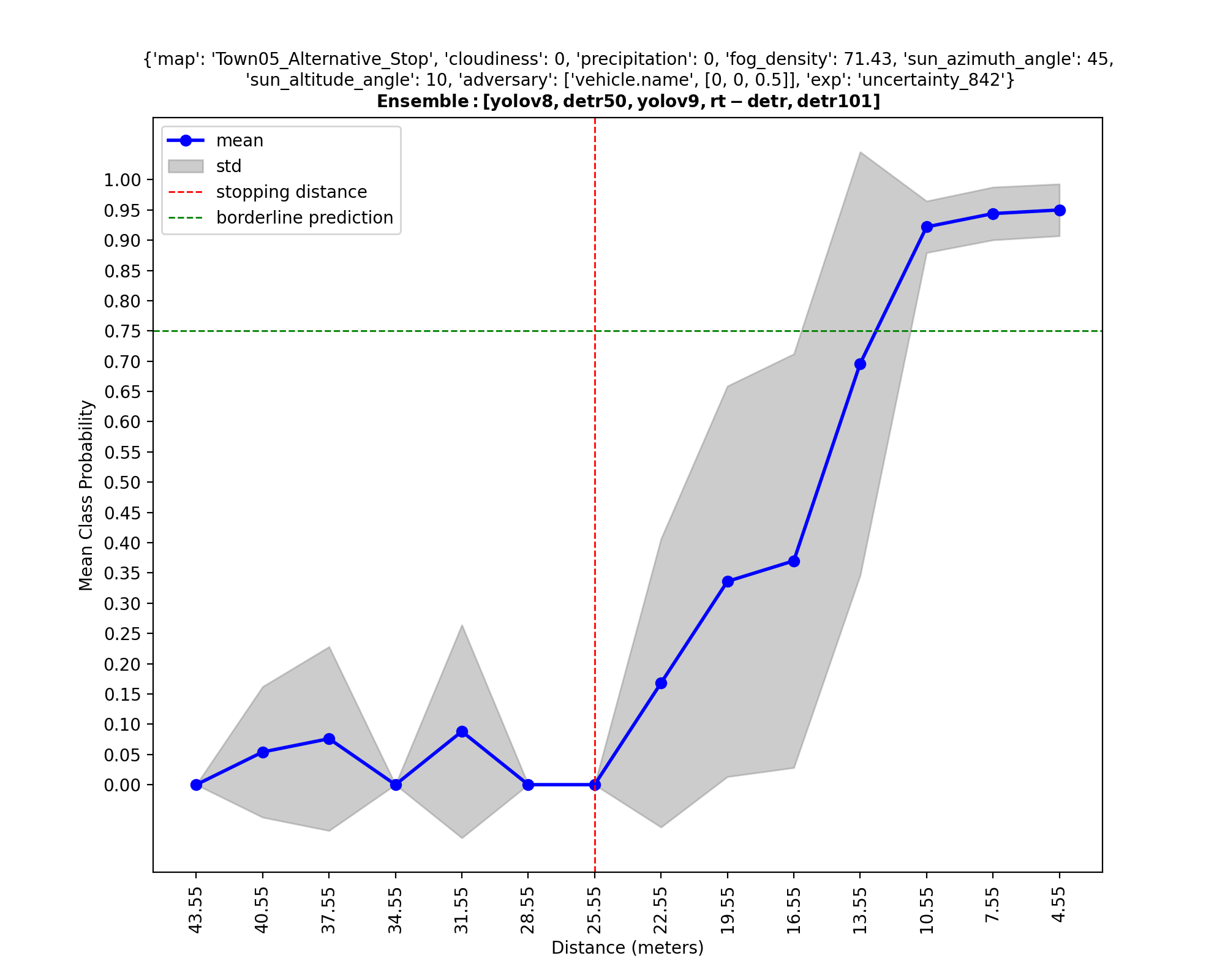}\label{fig:exp2_842g}}

    \caption{Case Study 4 (Experiment Set 2): Perception performance and visual context under adversarial stop sign alteration, high fog, and low light. This scenario demonstrates extreme robustness limits, profound predictive sensitivity, and a critical failure in detection, leading to an unsafe condition.}
    \label{fig:exp2_842_composite}
\end{figure*}

\subsubsection{Case Study 4: Extreme Robustness Limits with Altered Sign and Severe Fog/Low Light}
\label{sssec:results:exp2:case_study4}

Figure \ref{fig:exp2_842_composite} illustrates the perception performance in Experiment 842, a scenario demonstrating the extreme limits of the perception system's robustness under a combination of an adversarial object and severe environmental conditions. This experiment features an adversarially altered stop sign ($\mathcal{S}$) (contrast, texture, and color changed), combined with a high fog density (71.43\%) and a low sun altitude angle (10.0 degrees). These conditions (with 0\% cloudiness and precipitation) represent a profoundly challenging visual environment, as shown in Figure \ref{fig:exp2_842c}.

As depicted in Figure \ref{fig:exp2_842g}, the ensemble's mean prediction probability ($\mu_{\mathcal{S}}$) for $\mathcal{S}$ exhibits an alarming and persistent delay in detection. It remains at extremely low values, fluctuating slightly, for an extensive range (e.g., $\approx 0.02 \mbox{ at } d_{\mathcal{S}} = 43.55$ m, dropping to $\approx 0.02 \mbox{ at } 28.55$ m). Crucially, at the stopping distance ($sd = 25.55$ m), $\mu_{\mathcal{S}}$ is approximately 0.02, representing a clear and sustained failure to achieve the required confidence ($\theta_{\mathcal{S}} = 0.75$) to enter region ($\mathcal{R}$). The perception system remains outside $\mathcal{R}$ as the ego approaches $\mathcal{S}$. Confidence only begins to climb significantly at very close ranges, finally achieving the threshold around $d_{\mathcal{S}} \approx 11-12$ m, which is far beyond any safe reaction distance.

This profound delay in detection is accompanied by an standard deviation ($\sigma_{\mathcal{S}}$) band that is exceptionally wide and volatile throughout almost the entire detection range. This signifies pervasive model disagreement, indicating that models struggle not just with confidence, but also with basic agreement on object presence even at distances where high confidence is expected. The combined adverse effects of the altered sign, high fog, and low sun altitude critically degrade the perception system's reliability. This scenario unequivocally results in an unsafe driving condition, as the perception system provides utterly unreliable or delayed input within the necessary safety margins. It highlights operation far outside the defined ODD, underscoring extreme vulnerability to the combination of adversarial alteration and severe environmental obscuration.

The four case studies presented (Experiments 156, 325, 690, and 842) from Experiment Set 2 vividly demonstrate the perception system's robustness limits under intensified operating conditions. The introduction of extreme environmental parameters (e.g., high fog, high precipitation, severe low light) and complex adversarial alterations consistently led to profound increases in predictive sensitivity and severe detection degradation. These results highlight scenarios where the system either completely fails to achieve the required confidence within the stopping distance or exhibits highly unstable temporal behavior, leading to unsafe operating conditions. A wide standard deviation ($\sigma_{\mathcal{S}}$) band frequently served as a marker that participating models were unable to reach a consensus regarding detection and classification of the stop sign ($\mathcal{S}$). Furthermore, these experiments underscored that a low standard deviation, particularly when coupled with a low mean prediction probability ($\mu_{\mathcal{S}}$), is a critical indicator of systemic failure, pointing to a scenario where all participating models underperform and reach a consensus regarding their under-performance. The precise identification of these degradation thresholds and failure modes is essential for delineating the perception system's ODD, ensuring safe and reliable autonomous vehicle operation within its defined capabilities.

\subsection{Real-World Validation and Observed Predictive Sensitivity}
\label{subsec:results:real_world_validation}

The real-world test vehicle runs on a drive-by-wire kit provided by Dataspeed Inc.,\footnote{\url{https://www.dataspeedinc.com/}}. This real-world ego is equipped with  a LiDAR sensor, an image sensor (Lucid camera) and a radar sensor. For these experiments, only the images captured by the Lucid camera were utilized, with the predictive sensitivity quantification framework based on the architecture described in Figure \ref{fig:arch}. Profiling the ego occurred at two different speeds 10 MPH (Figure~\ref{fig:rw_10mph_composite}) and 25 MPH (Figure~\ref{fig:rw_25mph_composite}), as it approached the stop sign $\mathcal{S}$. In both scenarios, $\mathcal{S}$ is partially occluded by a person standing in front of it. Notably, human control was maintained for both real-world scenarios.

\subsubsection{Real-World Case Study 1: Confidence Reversal at 10 MPH}
\label{sssec:results:rw_case_study1}

\begin{figure*}[!ht]
    \centering
    \subfloat[Real-world camera view from 10 MPH scenario, showing stop sign partially occluded by a person at $d_{\mathcal{S}} = 9.0$ m. Note: Detected objects (person, car) are shown with bounding boxes and confidence scores from the perception system.]{\includegraphics[width=0.48\linewidth, valign=m, margin=0cm .9cm]{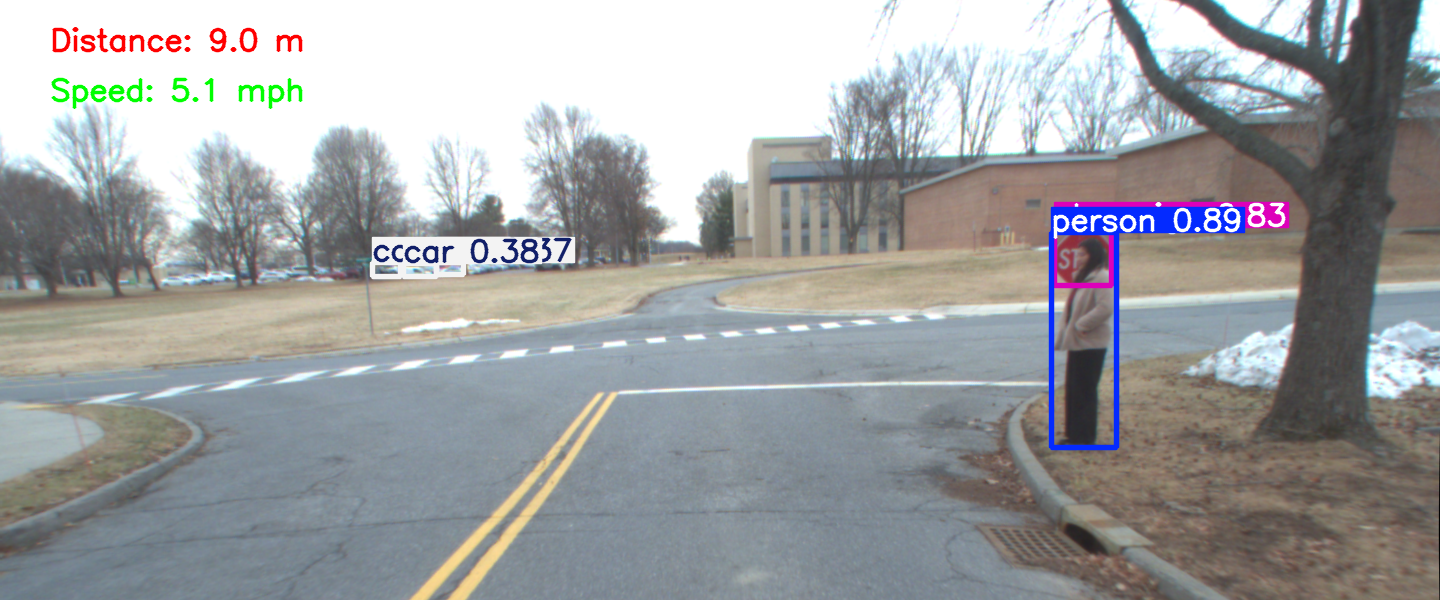}\label{fig:rw_10mph_c}}
    \hfill
    \subfloat[Mean prediction probability ($\mu_{\mathcal{S}}$) and standard deviation ($\sigma_{\mathcal{S}}$) for stop sign detection at 10 MPH. The vertical red dotted line indicates the stopping distance ($sd = 6$ m), and the horizontal green dotted line marks the confidence threshold ($\theta_{\mathcal{S}} = 0.75$). The gray shaded area represents $\sigma_{\mathcal{S}}$.]{\includegraphics[width=0.48\linewidth, valign=m]{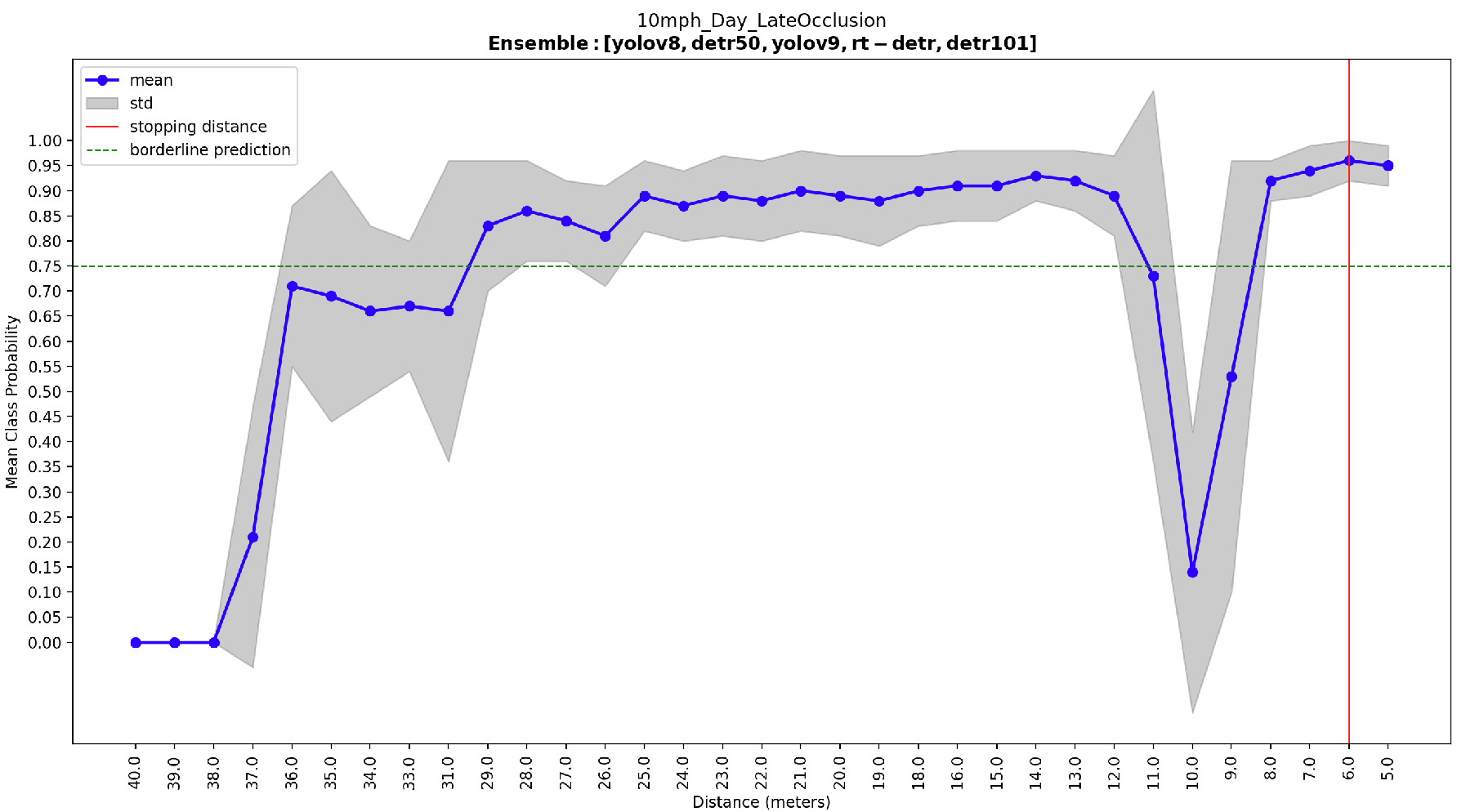}\label{fig:rw_10mph_g}}
    \caption{Real-World Case Study 1 (10 MPH): Perception performance and visual context under partial occlusion by a person. This scenario demonstrates a critical confidence reversal and loss of temporal consistency, despite initial high confidence, leading to an unsafe condition.}
    \label{fig:rw_10mph_composite}
\end{figure*}

Figure \ref{fig:rw_10mph_composite} presents the perception performance from a real-world experiment conducted at 10 MPH. In this scenario, the stop sign ($\mathcal{S}$) was partially occluded by a person, as captured by the Lucid camera (Figure \ref{fig:rw_10mph_c}). During the approach to $\mathcal{S}$, the human driver actively managed speed, slowing to around 5 MPH.

As depicted in Figure \ref{fig:rw_10mph_g}, the ensemble's mean prediction probability ($\mu_{\mathcal{S}}$) for $\mathcal{S}$ initially rises sharply from near zero at approximately $d_{\mathcal{S}} = 37$ m, achieving confidence above $\theta_{\mathcal{S}} = 0.75$ at around $d_{\mathcal{S}} = 28$ m. For a substantial range (from $\approx 28 \rightarrow 12$ m), $\mu_{\mathcal{S}}$ generally maintains this high confidence, and the standard deviation ($\sigma_{\mathcal{S}}$) band significantly narrows, indicating robust detection and low predictive sensitivity. However, a critical event occurs: at $d_{\mathcal{S}} \approx 10$ m $\mu_{\mathcal{S}}$ experiences a sharp and severe drop from $\approx 0.85$ to $\approx 0.15$, with a corresponding sharp rise in predictive  sensitivity. This abrupt reversal of confidence occurs precisely when the ego is within the critical operating range. $\mu_{\mathcal{S}}$ subsequently recovers to high confidence again as the ego approaches the stopping distance ($sd = 6$ m for 10 MPH), crossing $\theta_{\mathcal{S}}$ at $\approx 7$ m.

This pronounced drop in $\mu_{\mathcal{S}}$ represents a severe loss of temporal consistency within the safety quadrant ($\mathcal{R}$), as confidence unexpectedly plummets despite the vehicle getting closer to the object. The accompanying dramatic widening of the $\sigma_{\mathcal{S}}$ band (from $\approx 11 \rightarrow 9$ m) signifies extreme model disagreement during this confidence reversal. This erratic behavior constitutes an unsafe condition. Such a rapid, inexplicable loss of confidence could trigger erroneous downstream decisions (e.g., disengaging a braking maneuver), highlighting that perception instability, even if temporary, severely compromises safety. This real-world observation qualitatively validates the phenomena of fluctuating confidence and high predictive sensitivity previously observed in simulated scenarios, underscoring the relevance of our simulation findings, and demonstrates that even at low speeds, such an occlusion can cause the perception system to behave abruptly.

\subsubsection{Real-World Case Study 2: Systemic Failure at 25 MPH}
\label{sssec:results:rw_case_study2}

\begin{figure*}[!ht]
    \centering
    \subfloat[Real-world camera view from 25 MPH scenario, showing stop sign partially occluded by a person at $d_{\mathcal{S}} = 13.0$ m.]{\includegraphics[width=0.48\linewidth, valign=m, margin=0cm .9cm]{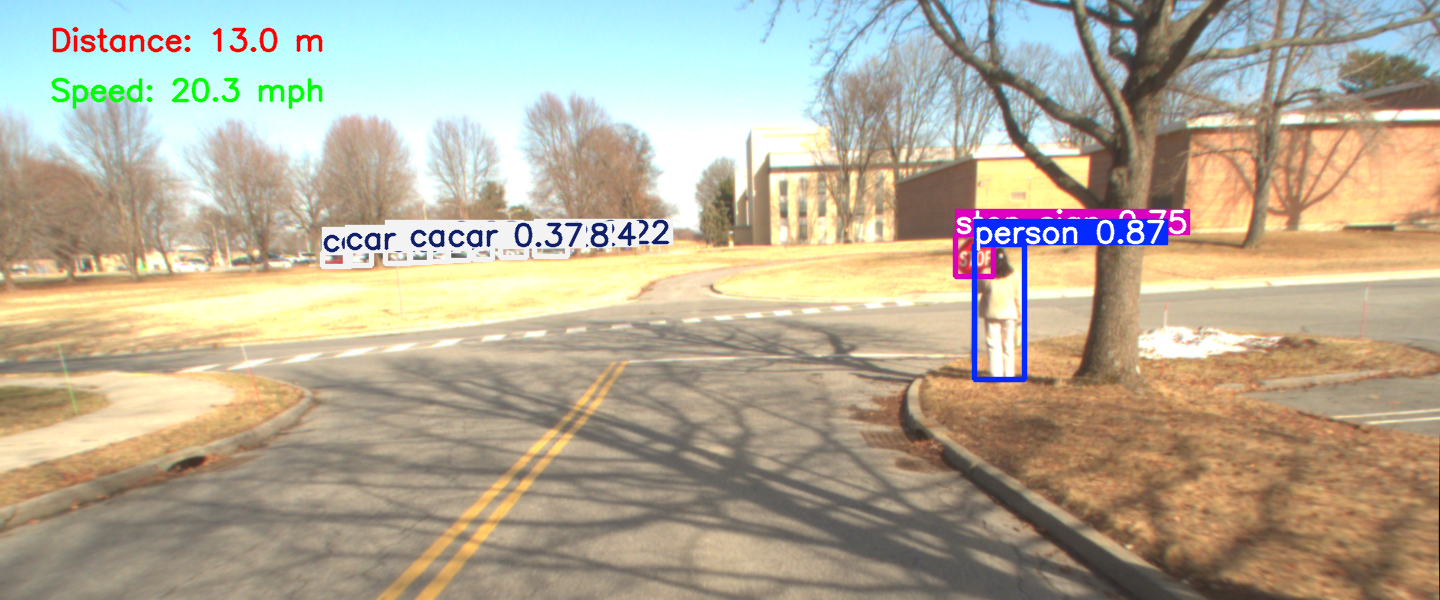}\label{fig:rw_25mph_c}}
    \hfill
    \subfloat[Mean prediction probability ($\mu_{\mathcal{S}}$) and standard deviation ($\sigma_{\mathcal{S}}$) for stop sign detection at 25 MPH. The vertical red dotted line indicates the stopping distance ($sd = 19$ m), and the horizontal green dotted line marks the confidence threshold ($\theta_{\mathcal{S}} = 0.75$). The gray shaded area represents $\sigma_{\mathcal{S}}$.]{\includegraphics[width=0.48\linewidth, valign=m]{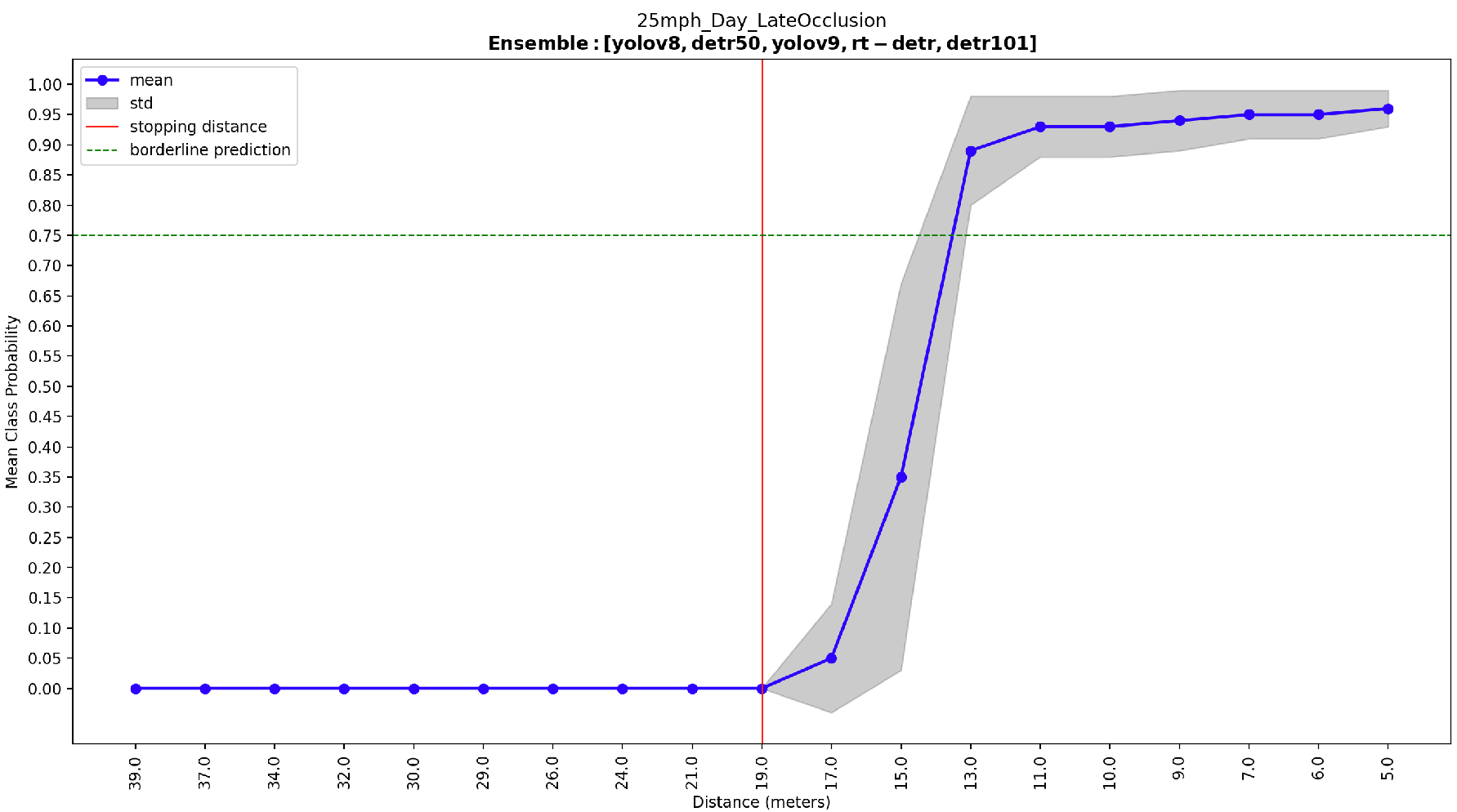}\label{fig:rw_25mph_g}}

    \caption{Real-World Case Study 2 (25 MPH): Perception performance and visual context under partial occlusion by a person. This scenario demonstrates a systemic failure in detection, leading to an unsafe condition.}
    \label{fig:rw_25mph_composite}
\end{figure*}

Figure \ref{fig:rw_25mph_composite} presents the perception performance from a real-world experiment conducted at 25 MPH. In this scenario, the stop sign ($\mathcal{S}$) was again partially occluded by a person, as captured by the Lucid camera (Figure \ref{fig:rw_25mph_c}).

As depicted in Figure \ref{fig:rw_25mph_g}, the ensemble's mean prediction probability ($\mu_{\mathcal{S}}$) for $\mathcal{S}$ consistently remains at or near zero for a substantial range from $d_{\mathcal{S}} = 39$ m until the stopping distance ($sd = 19$ m for 25 MPH). Crucially, at $sd = 19$ m, $\mu_{\mathcal{S}}$ is approximately 0.00. This complete lack of detection at the critical safety point is further exacerbated by a very narrow standard deviation ($\sigma_{\mathcal{S}}$) band around $\mu_{\mathcal{S}} = 0.00$ in this critical region. This indicates that all participating models were in consensus regarding the non-detection of $\mathcal{S}$, which represents a systemic failure for the perception system.

It is only after $sd$ that a sharp, monotonic increase in $\mu_{\mathcal{S}}$ is observed. The ensemble achieves confidence above $\theta_{\mathcal{S}} = 0.75$ just before $d_{\mathcal{S}} = 13$ m. While $\mu_{\mathcal{S}}$ rises sharply and remains high at closer distances, this significant delay in achieving confidence, coupled with the initial systemic failure to detect at $sd$, unequivocally pushes the ego vehicle into unsafe operating conditions. A similar trend of delayed detection and initial consensus on non-detection, indicative of systemic failure, was observed in some of the CARLA simulation case studies. This further establishes the hypothesis that the performance of perception systems is highly dependent on distance and various state variables.

The two real-world case studies presented (10 MPH and 25 MPH scenarios) effectively substantiate the perception sensitivity phenomena observed in the extensive simulation experiments. Despite the inherent logistical and physical limitations of real-world testing that restricted its scale, the observed behaviors -- including critical confidence reversals and instances of systemic failure (consensus on non-detection) --  demonstrate qualitative alignment with the diverse challenges and degradation patterns identified in both Experiment Set 1 and Experiment Set 2. This real-world evidence confirms that the simulation-based findings accurately reflect perception robustness in real-world contexts, providing confidence in the validity of the synthetic data for assessing autonomous vehicle performance.

\section{Discussion}
\label{sec:discussion}

The comprehensive experimental evaluation, encompassing both extensive simulation scenarios and targeted real-world validation, provides critical insights into the robustness and sensitivity of perception systems in autonomous vehicles. This work systematically identified key environmental and adversarial conditions that profoundly impact perception performance, leading significant degradation in detection. 

The results demonstrate that factors such as high fog density, heavy precipitation, low sun altitude (causing low light or glare), and adversarial alterations or occlusions consistently push perception systems to their operational limits, often resulting in unsafe conditions. Furthermore, the qualitative alignment between simulated and real-world observations underscores the practical relevance of these findings for enhancing autonomous vehicle safety.

These findings carry significant implications for the safe deployment of autonomous vehicles, particularly in defining their ODDs. The identification of specific environmental and adversarial conditions that reliably induce catastrophic perception failures provides crucial data for delineating these safe operating boundaries. The observed behaviors, such as significant delays in detection, abrupt confidence reversals, and instances of systemic failure (where models agree on non-detection in critical scenarios), directly compromise the perception system's ability to provide the required input for safe decision-making. Furthermore, the critical importance of evaluating temporal consistency in perception outputs is underscored, as even transient losses of confidence can render the system unreliable. While a robust perception system is a necessary condition for autonomous vehicle safety, these results highlight that achieving high confidence from the perception module alone does not guarantee overall safety, underscoring the need for robust integration with planning and control systems.

\subsection{Limitations of the Study}
\label{subsec:discussion:limitations}

Despite its significant contributions, this study has several limitations. The experimental scope primarily focused on the detection of a single object class (stop signs) within specific simulated and real-world driving scenarios. In designing the synthetic data experiments (outlined in Sec. \ref{sec:methodology}), limitations imposed by the chosen simulator, CARLA, became apparent, influencing options for introducing uncertainty. CARLA does not offer photorealistic rendering, which can increase the domain shift between real-world and synthetic images. This concern is particularly relevant for weather effects, which are notably simplified in simulation. Moreover, real-world cameras at higher speeds exhibit motion blurring effects, unlike CARLA's cameras, which provide comparatively clear images even at high speeds.

A more fundamental limitation of the experiment design was the choice to test all combinations of weather and adversary parameters in order to assess their combined effects on sensitivity assessment. This caused a combinatorial explosion in the number of scenarios to consider, particularly as more parameters were included. This approach, while comprehensive for a limited set of parameters, becomes computationally intractable for broader parameter space exploration. This prevented testing a wider range of potential factors or discretizing the tested factors more finely. The most important case where this is relevant is when multiple factors interact to create complex occlusions or distortions.

Furthermore, the qualitative nature of the real-world validation, while crucial for corroborating simulation findings, limits quantitative generalization. Additionally, the stopping distance calculations, while practical for assessing perception system output, utilized a simplified physics model that does not account for complex vehicle dynamics or passenger comfort parameters. These limitations narrow the immediate generalizability of the absolute performance metrics; however, they do not detract from the qualitative understanding of perception sensitivity and the methodology for its assessment, which remains broadly applicable.



\subsection{Directions for Future Work}
\label{subsec:discussion:future_work}

Building upon the insights gathered during this study, several promising directions for future research emerge to enhance methodology for assessing the robustness of autonomous vehicle perception systems. Potential enhancement may include considering addition AV operating behaviors beyond signage detection and response. Also, exploring sensitivity under extrinsic conditions to better understand perception sensitivity to factors such as speed. This includes investigating alternative methods, such as generative AI for producing photorealistic  data, particularly for challenging weather effects and motion blur.

Avoiding the combinatorial explosion observed in parameter space exploration can be achieved by relying on more efficient random sampling techniques. Alternatively, search of the parameter space, informed by high-sensivity scenarios, can be used.

\section{Conclusion} \label{sec:conclusion}

This study investigated sensitivity assessment of autonomous vehicle perception systems through robustness evaluation under various environmental and adversarial conditions. Extensive simulation experiments and supporting real-world validation identified critical factors that impact perception performance, leading to increased ensemble standard deviation and significant degradation in detection. The findings underscore the importance of evaluating factors such as latency and the risk of systemic failures (non-detections or phantom detections), which directly compromise safe operation. This research also delineates severe limits of perception robustness and provides direction for extending the definition of the ODD of an autonomous vehicle, thereby enhancing their safety and reliability in complex real-world environments.

\bibliographystyle{IEEEtran}
\bibliography{references}

\flushend 
\end{document}